%% file: main.tex
\documentclass[10pt,twocolumn,letterpaper]{article}

\usepackage{cvpr}              

\usepackage[accsupp]{axessibility}   
\usepackage{graphicx}
\usepackage{amsmath}
\usepackage{amssymb}
\usepackage{booktabs}
\usepackage{xspace}
\usepackage{maths} 

\def\R{{\mathbb{R}}} 
\usepackage[pagebackref,breaklinks,colorlinks]{hyperref}
\usepackage{pifont}
\usepackage{xcolor}
\usepackage{textcomp}
\usepackage{dsfont}
\usepackage{multirow}
\usepackage{enumitem}
\usepackage{rotating}

\def\mypar#1{\vspace{0mm}{\noindent\bf #1.}\hspace{1mm}}
\def\fig#1{Figure~\ref{fig:#1}}
\def\tab#1{Table~\ref{tab:#1}}

\def\Eq#1{Eq.~(\ref{eq:#1})}

\def\myvspace{\vspace{-1mm}} 

\makeatletter
\DeclareRobustCommand\onedot{\futurelet\@let@token\@onedot}
\def\@onedot{\ifx\@let@token.\else.\null\fi\xspace}

\def\eg{\emph{e.g}\onedot} 
\def\ie{\emph{i.e}\onedot}

\def\vs{\emph{vs}\onedot}
\def\wrt{w.r.t\onedot} 
\def\etal{\emph{et al}\onedot}
\makeatother

\newcommand{\cmark}{\textcolor{green!80!black}{\ding{51}}}
\newcommand{\xmark}{\textcolor{red}{\ding{55}}}

\def \ours {{CAT}\xspace}
\def \source {{\mathcal{S}}\xspace}
\def \target {{\mathcal{T}}\xspace}

\usepackage[capitalize]{cleveref}
\crefname{section}{Sec.}{Secs.}
\Crefname{section}{Section}{Sections}
\Crefname{table}{Table}{Tables}
\crefname{table}{Tab.}{Tabs.}

\def\cvprPaperID{9115} 
\def\confName{CVPR}
\def\confYear{2023}

\begin{document}

\title{Few-shot Semantic Image Synthesis with Class Affinity Transfer}

\author{
Marl\`ene Careil$^{1,2}$  \quad
Jakob Verbeek$^{2}$  \quad
St\'ephane Lathuili\`ere$^{1}$
\\
$^1$LTCI, T\'el\'ecom Paris, IP Paris\quad
$^2$Meta AI
}

\maketitle


 \input{abstract}
 \input{intro}

 \input{related}
 \input{method}
 \input{exp}
 \input{conc}

\section*{Supplementary material} \input{supmat}

\clearpage
{\small
\bibliographystyle{ieee_fullname}
\bibliography{main}
}

\end{document}

%% file: abstract.tex
\begin{abstract}

Semantic image synthesis  aims to generate photo realistic images given a semantic segmentation map. 
Despite much recent progress, training them still requires large datasets of images annotated with per-pixel label maps that are extremely tedious to obtain. 
To alleviate the high annotation cost, we propose a transfer method that leverages a  model trained on a large source dataset to  improve the learning ability on small target datasets via estimated pairwise relations between source and target classes. 
The  class affinity matrix is introduced as a first layer to the source model to make it compatible with the target label maps, and   the source model is then further finetuned for the target domain.
To estimate the class affinities we consider different approaches to leverage  prior knowledge:  semantic segmentation on the source domain, textual label embeddings, and  self-supervised vision features.
We apply our approach to GAN-based and diffusion-based  architectures for semantic synthesis. 
Our experiments show that the different ways to estimate class affinity can be effectively combined, and that our approach significantly improves over existing state-of-the-art transfer approaches for generative image models. 

\end{abstract}

%% file: intro.tex
\section{Introduction}

Image synthesis with deep generative models has made remarkable progress in the last decade with the introduction of GANs~\cite{goodfellow14nips}, VAEs~\cite{kingma14iclr},  and diffusion models~\cite{ho20neurips}. 
Generated images can be conditioned on diverse types of inputs, such as class labels~\cite{brock19iclr,mirza14nips}, text~\cite{gafni22arxiv,reed16icml,ramesh22dalle2}, bounding boxes~\cite{sun19iccv2}, or seed images~\cite{casanova21nips}. 
In semantic image synthesis, the generation is conditioned on a semantic map that indicates the desired class label for every pixel. 
This task has been thoroughly explored with models such as SPADE~\cite{park19cvpr1} and OASIS~\cite{sushko21iclr}, capable of generating high-quality and diverse images on complex datasets such as ADE20K~\cite{zhou17cvpr}
and COCO-Stuff~\cite{caesar18cvpr}.
However, these approaches heavily rely on the availability of large datasets with tens to hundreds of thousands of images annotated with pixel-precise label maps that are extremely costly to acquire. 
For the Cityscapes dataset~\cite{cordts16cvpr}, \eg,  on average more than 1.5h per image was required for annotation and quality control.

\begin{figure}
 \def\myim#1{\includegraphics[width=27mm,height=27mm]{samples/#1}}
     \centering
   \setlength\tabcolsep{1.5 pt}
   \small
\begin{tabular}{ccc}
\multirow{2}{*}{Input segmentation} & Class affinity transfer,  & Standard training,  \\
&  100 training images & 20k training images  \\
\myim{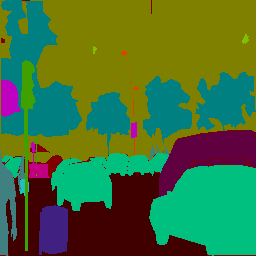} &
\myim{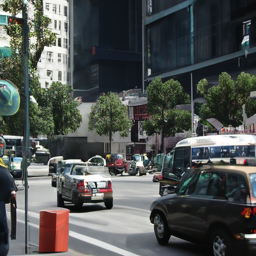} &
\myim{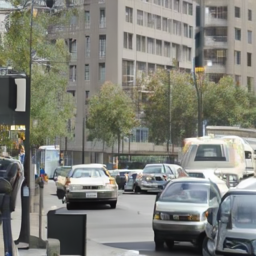} 
\\
\myim{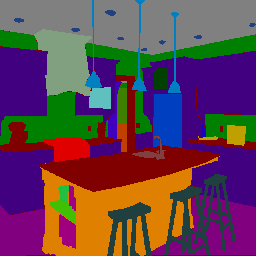} &
\myim{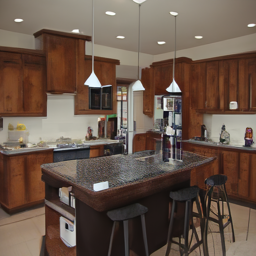} &
\myim{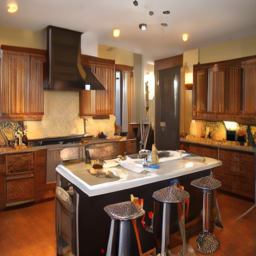} 
\end{tabular}
\myvspace
\caption{ 
{\bf Can we train a semantic image synthesis model from only 100 images?} 
Our diffusion-based transfer results using training set of 100 ADE20K images (2$^\textrm{nd}$ col.) compared to the same model trained from scratch on full dataset (20k images, 3$^\textrm{rd}$ col.).
}
\label{fig:teaser}
\end{figure}

High annotation costs can be  a barrier to deployment of machine learning models in practice, and motivates 
the development of transfer learning strategies to alleviate the annotation requirements. 
These techniques allow  training models  on small target datasets via the use of models pre-trained on a source dataset with many available annotations. 
Transfer learning has been widely studied for classification tasks such as object recognition~\cite{rebuffi17nips,berriel19iccv,jia2022visual}, but received much less attention in the case of generation tasks. 
This task has been considered for unconditional and class-conditional generative models~\cite{mo20cvpr,noguchi19iccv,ojha21cvpr,wang18eccv2,wang20cvpr,zhao20icml}, but to the best of our knowledge few-shot transfer learning has not yet been explored in the setting of semantic image synthesis.

We introduce \textbf{\ours}, a finetuning procedure that models \textbf{C}lass \textbf{A}ffinity to \textbf{T}ransfer knowledge from pre-trained semantic image synthesis models. 
Our method takes advantage of prior knowledge to establish pairwise relations  between  source and target classes, and encodes them in a class affinity matrix.
This solution considerably eases learning when few instances of the target classes are available at training time. 
The affinity matrix is prepended to the source model to make it compatible with the label space of the target domain.  
The model can then be further finetuned using the available data for the target domain. 
To illustrate the generality of the proposed approach, we integrate our transfer learning strategy in  state-of-the-art  adversarial  and diffusion  models. 
We explore different ways to extract similarities between source and target classes, using semantic segmentation models for the source data, self-supervised vision features, and text-based class embeddings.

We  conduct extensive  experiments on the ADE20K, COCO-Stuff, and Cityscapes datasets, using  target datasets  with sizes ranging from as little as 25 up to 400 images.
Our experiments show that our approach significantly  improves over state-of-the-art  transfer methods. 
As illustrated in \fig{teaser}, our approach allows realistic synthesis from no more than 100 target images, and achieves image quality close to standard training on the full target datasets.  
Moreover, unlike previous transfer methods, our approach also enables non-trivial training-free transfer results, where we only prepend the class affinity matrix to the source model, without further finetuning it. 

\noindent 
In summary, our contributions are the following:
\begin{itemize}[noitemsep,topsep=0pt]
\item We introduce Class Affinity Transfer (CAT), the first transfer method for  semantic image synthesis for  small target datasets, and explore different methods to define class affinity, based on semantic segmentation, self-supervised features, and text-based similarity. 
\item We integrate our approach in state-of-the-art 
adversarial and diffusion based  semantic  synthesis models.  
\item We obtain excellent experimental transfer results, improving over existing state-of-the-art approaches.
\end{itemize}

%% file: related.tex
\section{Related work}

\mypar{Semantic image synthesis with GANs}
There has  been significant interest in adversarial approaches for semantic image synthesis, see \eg~\cite{azadi20arxiv,park19cvpr1, isola17cvpr,sushko21iclr}. These approaches employ a conditional generator and a discriminator that assesses both image quality and consistency with the input segmentation maps. One of the first models proposed was Pix2Pix\cite{isola17cvpr}, which uses a U-Net~\cite{ronneberger15miccai} generator along with a patch-based discriminator.  SPADE~\cite{park19cvpr1} employs a different generator with spatially adaptive normalization layers modulating feature maps through labels. Lab2Pix-V2\cite{lab2pix} introduces special modules in the generator for extracting meaningful information from labels.  OASIS~\cite{sushko21iclr} overcomes the need of perceptual loss with the introduction of a U-Net discriminator which produces per-pixel classification scores.  This approach obtains state-of-the-art results on the task of semantic image synthesis, and we build upon it 
in our GAN-based experiments.

\mypar{Semantic  synthesis with diffusion-based models}
Recently,  diffusion models~\cite{ho20neurips} have emerged as a promising solution capable of synthesizing images with a quality that surpasses GANs~\cite{dhariwal21nips,rombach21arxiv}. Generation is formulated as an iterative denoising process and a likelihood-based loss is used as training objective which makes training more stable and scalable to large datasets. 
A few works address semantic image synthesis with diffusion models. In \cite{wang2022semantic}, SPADE blocks are included in the U-Net used in the diffusion model to improve the semantic consistency of the generated images. PITI~\cite{wang22arxiv}  builds upon GLIDE~\cite{nichol21arxiv}, a  text-conditioned diffusion model pre-trained on hundreds of millions of image-text pairs. The text encoder is then replaced by one that takes semantic segmentation maps as input. 
Different from our work, PITI focuses on transferring a text-based model to a semantic synthesis model, and still requires a large training set  to train the semantic map encoder network from scratch (20k to 110k training images in their experiments on ADE20K and COCO-Stuff). 
In contrast, we address transfer from existing semantic synthesis models, in  scenarios where only few target images are available: from 25 to 400 in our experiments. 

\begin{figure*}[t]
\centering
\includegraphics[width=0.85\textwidth]{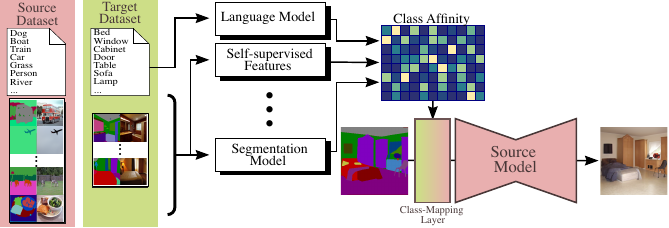}
\myvspace
\caption{
Overview of our class affinity transfer (CAT) approach for semantic image synthesis.
The class affinity matrix is used to align the source model with the target label space, and is then further finetuned using the  target images and corresponding segmentations.    
}
\label{fig:pipeline}
\end{figure*}

\mypar{Transfer learning for generative models}
 Compared to transfer learning for discriminative problems, transfer for generative models received much less attention.
In the seminal work~\cite{wang18eccv2},  a pre-trained unconditional GAN is finetuned either for conditional or unconditional generation in limited data regimes. 
Several works show that this approach can be improved by finetuning only part of the network parameters~\cite{noguchi19iccv,mo20cvpr}.
Another strategy consists in adding and  learning a limited number of  additional parameters~\cite{robb20arxiv,wang20cvpr,zhao20icml}. For instance, Zhao \etal~\cite{zhao20icml}
apply affine modulations to the frozen parameters of the pre-trained model. 
In MineGAN~\cite{wang20cvpr}, a small ``miner'' network is introduced to warp the distribution of the latent variable to better fit the target distribution. 
In ~\cite{Gu_2021_ICCV}, they perform few-shot image generation by using a local fusion module in the encoder space. Other approaches investigate the use of style transfer for few-shot GAN training~\cite{ojha21cvpr, zhao2022closer}. 

A few  works specifically consider transfer for conditional GANs.
Shahbazi \etal~\cite{shahbazi21cvpr} propagate information from old classes to new classes through the use of batch normalization. 
Their work, however, considers class conditional generation, which is different from our work where we condition on semantic segmentation maps.
Endo \etal~\cite{endo21arxiv} use an unconditional StyleGAN2 model to generate synthetic images with  pseudo labels.
They do this by learning a nearest-centroid classifier in the GAN latent space from real images with corresponding label maps, where a PSP encoder~\cite{richardson21cvpr} is used to obtain the latents of real images. 
Although interesting, this approach hinges on the availability of a strong unconditional generative model and the ability of faithful latent inference, which is possible for datasets with limited diversity such as human faces, but which is extremely challenging for complex datasets such as ADE20K and COCO-Stuff that we consider in our experiments.

There are very few works on model transfer for diffusion models.
In addition to PITI, which considers semantic image synthesis as discussed above, Ruiz \etal~\cite{ruiz2022dreambooth} consider the problem of instance-driven generation. They finetune a pretrained text-based diffusion model on a handful of images of a particular object in different contexts. The finetuned model is then  used to  generate  images with the same object in other environments described by a textual prompt. 
 

%% file: method.tex
\section{Class affinity transfer}
\label{sec:method}

We aim at adapting a semantic image synthesis model pre-trained on a large source dataset to a small target dataset. 
We assume a source dataset composed of $N$ RGB images $ \Xmat_n\!\in\!\R^{H\times W\times 3}, n\!\in\!\{1,\dots,N\}$ and their corresponding segmentation maps $\Smat_n\!\in\!\{0,1\}^{H\times W\times C_\source}$, represented with one-hot encoding across $C_\source$  source classes. Similarly, we consider a target dataset composed of $M$ images $\Xmat_m^\target\!\in\!\R^{H\times W\times 3}, m\in\{1,\dots,M\}$ and their corresponding segmentation maps $\Smat_m^\target\!\in\!\{0,1\}^{H\times W\times C_\target}$ with $C_\target$ target classes. Note that, the number of classes is different between the source and target datasets, and that no correspondence between them is given.

Our class affinity transfer (\ours) approach is based on an affinity matrix $\Amat\in\R^{C_\source\times C_\target}$ that maps between the classes of the source and target datasets. 
We use the affinity matrix to parameterize a linear layer which we 
prepend  to the source model, making the source model compatible with one-hot input label maps of the target domain. 

Rather than initializing the affinity matrix at random, we leverage different forms of prior knowledge to estimate the affinity matrix, which enables the source model to adapt significantly better to the target domain. 
The model can then be further finetuned using training images and segmentation maps from the target domain. 
In addition, our approach also allows for ``training-free'' transfer mode where we fully rely on the class affinity transfer matrix, further  finetuning on the target dataset.

Below, we  describe different approaches to estimate the class affinity matrix in  \S\ref{sec:classsimextr}. 
Then, we show how our approach can be incorporated into a state-of-the-art GAN and diffusion based architectures in \S\ref{sec:archi} and  \S\ref{sec:meth_diffusion}, respectively.

\subsection{Estimating the class affinity matrix}
\label{sec:classsimextr}

We consider three different ways to estimate the class affinity  matrix $\Amat$ between the source and target classes, 
leveraging different forms of prior knowledge.

\mypar{Supervised semantic segmentation networks}
Here we employ a pre-trained segmentation network trained on the source dataset to extract mappings between source and target classes.
First, we segment the target images across the source classes using the segmentation network. 
Next, we use segmentation maps of the target images and count how many pixels of each target class are classified as each source class to establish the class affinity matrix. 

More formally, we denote by $\bar{\Smat}_m\in \{0,1\} ^{ H\times W \times C_\source}$ the output of the segmentation network for a target image $\Xmat_m^\target$, and use  subscripts $i,j$ to denote pixel locations, and superscripts $k$ and $l$ to index across target and source classes. 
The  affinity matrix $\Amat$ is computed as the confusion matrix between source and target classes:
\begin{equation}
   \Amat_{k,l}  \propto  \sum_{m=1}^{M} \sum_{i,j} 
[\Smat_m^\target]_{i,j}^k \cdot [\bar{\Smat}_m]_{i,j}^l.
    \label{eq:CondDiscLoss}
\end{equation}
The matrix  is normalized so that for each  target class the affinities \wrt  all source classes sum to  one, \ie $ \sum_{l=1}^{C_\source} \Amat_{k,l} = 1$.
In this manner, prepending the affinity matrix as  a linear layer to the network leaves the scale of the input comparable with inputs from the source dataset.
In our experiments, we use UperNet~\cite{xiao18eccv} or DeeplabV2~\cite{chen15iclr} as pretrained segmentation networks.

\mypar{Self-supervised image features}
To alleviate the requirement of training a dedicated segmentation network on the source dataset, we explore self-supervised learning (SSL) to extract features from image patches using iBOT~\cite{zhou21iclr}. 
Using the corresponding segmentation maps,  we represent each class in the source and target dataset using a ``prototype'' which is obtained as the   weighted average of the  features of patches that belong to that class. 
The features of each patch are weighted proportionally to the number of pixels with a given label in the patch. 
We denote these prototypes as  $\fvect_l^\source \in \R ^{D}$ and $\fvect_k^\target \in \R ^{D}$ where D is the embedding dimension, for source and target classes respectively. 
We then compute the class affinities  using cosine similarity between the prototypes:
\begin{eqnarray}
     \Amat_{k,l} \propto \cos( \fvect_k^\target , \fvect_l^\source),
    \label{eq:AfMat}
\end{eqnarray}
and similarly normalize the affinities so that for each target class  they sum to one across the source classes. 

\mypar{Text-based class affinities}
The previous approaches estimate class affinities using source and target images with corresponding segmentation maps. 
Here we consider an alternative that does not require any labeled images, and instead  relies on the class names to establish affinities. 
To this end,  we use a pre-trained CLIP~\cite{radford21clip} text encoder to embed the names of the source and target classes as  $\fvect_l^\source \in \R ^{D}$ and $\fvect_k^\target \in \R ^{D}$.
Similar to \Eq{AfMat}, we obtain the affinities as  normalized cosine similarities over the  text embeddings.

\mypar{Combination via majority voting} To take advantage of the three different methods to estimate the class affinities, we introduce an aggregation scheme that combines the affinity matrices obtained with all the previous methods.  
While the different estimations of $\Amat$ could be combined via simple averaging, we obtain better performance with a binary majority voting scheme. 
If, for a given target class, at least two of the three affinity matrices agree on source class with highest affinity,  then the target class is associated with the corresponding source class. If the three affinity matrices disagree, we take the source class provided by the method with the lowest initial, \ie ``training free'', FID on the training target set.

\subsection{Few-shot transfer with GAN}
\label{sec:archi}

We integrate our class affinity transfer approach with  the state-of-the-art OASIS semantic image synthesis model~\cite{sushko21iclr}.
It consists of a convolutional generator with SPADE blocks~\cite{park19cvpr1} to condition on segmentation maps, and a U-Net~\cite{ronneberger15miccai} discriminator to label pixels of real images with the corresponding class, and generated pixels as ``fake''. 
Based on initial experiments, we introduce several modifications to both generator and discriminator  to improve transfer to small target datasets.

\mypar{Architecture} First, we prepend the class affinity matrix to the SPADE blocks that take the segmentation map as input to align them with the target label space. 
Second, rather than sharing the first convolutional layer, we use separate paths for the  scale and shift parameters, but still share the parameters of the first convolutional layer. 
Third, we add two parallel branches from the input which  bypass the class affinity matrix and  the first  convolutional layer. 
These branches take  the target segmentation map as input and project the latter to the feature space of the first pretrained convolution output in each of the SPADE block. 
We then sum these residual outputs to the main branch.
The weights of the parallel branch are initialized with zeros to prevent negative impact early in training.
The motivation behind this design choice is to enable the generator to better learn how to synthesize new target classes which could not be explained by a linear combination of source classes. 

In the discriminator, we replace the last convolutional layer (which outputs per-pixel classification scores)  by a randomly initialized layer with an  output channel size corresponding to the number of classes in the target dataset. 
Alternatively, we experimented with a linear layer initialized with our affinity matrix added on top of the discriminator to map the source and target classes, but this approach did not improve performance.

\mypar{Finetuning}
Similar to~\cite{mo20cvpr}, we found that freezing the first layers of the discriminator is beneficial when finetuning the source model for the target datasets. 
Regarding the generator, we proceed in two stages. 
In the first stage, we fix most generator parameters and  only finetune the  class affinity matrix, the following first convolution layer, and the residual branch in each SPADE block. In the second stage, we finetune all the layers in the generator.
The losses used during finetuning are the same as during pretraining, \ie we use an adversarial loss as well as the LabelMix~\cite{sushko21iclr} regularization loss.

\subsection{Few-shot transfer with diffusion model}
\label{sec:meth_diffusion}

\mypar{Architecture}
For our diffusion-based experiments, we
 adopt the PITI~\cite{wang22arxiv}.
It is a modified version of GLIDE~\cite{nichol21arxiv}, a text-conditioned diffusion  model that generates the image via iterative denoising using a  U-Net.  
GLIDE consists of two text-conditional networks: the first generates a $64\!\times\!64$ image;  the second upsamples the image to $256\!\times\!256$ resolution.
In PITI,  the text encoder of both networks is replaced by a semantic map encoder with a transformer architecture.

To be compatible with our class affinity transfer approach, we modify PITI to be conditioned on one-hot label maps rather than RGB label maps. 
We do this by factoring the class-to-RGB mapping into the weights
of the first layer of the encoder network. 
Similarly to the GAN-based model, we use the class affinity matrix $\Amat$ to parameterize a linear layer which we prepend to the semantic image encoder. 
To allow further adaptation to the target task, we take inspiration from~\cite{jia2022visual}, and introduce trainable extra parameters in the transformer encoder, referred to as ``prompts'', which can be seen as additional  patch embeddings in input of each attention layer.
The prompts are randomly initialized. 
For more details see the supplementary material.

\mypar{finetuning} 
To finetune PITI, we freeze the decoder layers, \ie the U-Net model, and only train part of the segmentation encoder. 
We fix all the weights in the encoder transformer, and only train the last ResNet block and the prompts of the encoder. 
We employ the training loss used in GLIDE, and  
 finetune both the low resolution model as well as the conditional upsampling model, as in \cite{wang22arxiv}. 

%% file: exp.tex
\section{Experiments}
\label{sec:exp}

\subsection{Experimental setup}
\label{sec:setup}

\mypar{Datasets}
In order to make our research results comparable to earlier work on semantic image synthesis, we employ ADE20K~\cite{zhou17cvpr} (20k images and 151 classes) and COCO-Stuff~\cite{caesar18cvpr} (110k images and 183 classes) as source datasets. 
As target datasets, we use subsets of ADE20K and COCO-Stuff, as well as the  Cityscapes~\cite{cordts16cvpr} dataset consisting of 3k images and 35 classes.

To avoid training our models on personal data, we use the version of the Cityscapes dataset with blurred faces and license plates, while for COCO-Stuff and ADE20K we applied a face blurring pipeline ourselves.
Following~\cite{park19cvpr1,sushko21iclr}, we train  models at $256\!\times\!256$ resolutions for ADE20K and COCO-Stuff, and $256\!\times\!512$ for Cityscapes. 
For PITI~\cite{wang22arxiv},  we also use  a resolution of $256\!\times\!256$ for  Cityscapes since the  positional embeddings in PITI are pre-trained at $256\!\times\!256$ .

We sample subsets as target datasets to evaluate the different methods in few-shot regimes. To ensure that all the target classes are well represented in the subsets, we use a specific sampling procedure. 
We take an initial random image and then iteratively select subsequent images such that  the KL-divergence between the uniform distribution and the empirical class  distribution is minimized. 
The empirical class distribution is obtained by counting how many pixels of each class are present in each segmentation map, and normalizing the histogram to sum to one.
Unless otherwise indicated, we use target subsets of 100 images in all our experiments. The impact of the target set size is discussed in Section~\ref{sec:setsize} where we perform experiments with subsets  ranging in size from 25 to 400 images.

\mypar{Evaluation metrics}
We report both FID~\cite{heusel17nips} and mIoU metrics as in~\cite{sushko21iclr,park19cvpr1,isola17cvpr}. 
FID captures both image quality and diversity, while mIoU assesses the semantic correspondence with the input segmentation maps by using a segmentation network to label generated images. We use the same segmentation networks as in \cite{sushko21iclr}.

\mypar{Baselines}
For OASIS~\cite{sushko21iclr}, the most basic  comparison is to training 
the model from scratch, without any transfer.
To the best of our knowledge, we are the first to propose a transfer method specifically developed for semantic image synthesis.
Therefore, to evaluate our model, we compare to existing transfer learning works developed for unconditional and class-conditioned GANs by adapting them for semantic image synthesis. 
 We compare to TransferGAN~\cite{wang18eccv2} by finetuning all the layers of the source generator and discriminator to adapt to the target dataset. 
We also compare to Freeze-D~\cite{mo20cvpr}, which finetunes both generator and discriminator, while freezing the  layers of the discriminator closest to the input image. 
Based on the ablations in~\cite{mo20cvpr}, we consider freezing the first up to the ninth layers of the discriminator. 
BSA~\cite{noguchi19iccv} finetunes only the batch normalization (BN) parameters. In OASIS, the BN parameters are computed from the label maps through SPADE blocks. Therefore, for BSA we only finetune the first layer of the SPADE blocks and freeze all the other weights. 
MineGAN~\cite{wang20cvpr} freezes the source generator and adds a small MLP mapping network that transforms the latent vector, while finetuning the source discriminator to the target dataset. 
To adapt it to  semantic image synthesis, we also learn the class embedding layer of the generator for the target dataset.  
We follow the  two-stage training approach of MineGAN, where in the second training stage we finetune  the entire generator and discriminator networks. 
We also test cGANTransfer~\cite{shahbazi21cvpr}, by finetuning scale and shift parameters in SPADE blocks, projecting each target class as a trainable linear combinations of source classes. Unlike \ours, these linear combinations are initialized randomly. We also add trainable residual layers after the first convolution projecting label maps and train with the  $\ell_1$ and $\ell_2$  regularization losses of~\cite{shahbazi21cvpr}.

\input{fig3}

For PITI~\cite{wang22arxiv}, we compare our method to finetuning from a pretrained GLIDE~\cite{nichol21arxiv} model. In this baseline, we train the encoder from scratch to map segmentation maps to the latent space of GLIDE,  as in~\cite{wang22arxiv}. We also consider a baseline where we finetune all layers in a pretrained PITI, by re-initializing the first encoder layer that takes as input the segmentation map.

\input{table1tall}

\subsection{Main affinity estimation results}

\input{tab2.tex}

\mypar{Quantitative evaluation}
In our first experiment we evaluate the performance of the class affinity estimation from class label embeddings, semantic segmentation, self-supervised features, and their combination, when transferring from COCO-Stuff to ADE20K and vice-versa.
We also compare  to randomly initializing the affinity matrix. In all cases, we  initialize the other weights from the source model, and finetune all weights on the target data.

The results in \tab{effectClsSimExtrc} show consistent gains over the random initialization baseline across the board. 
The text and segmentation based affinities perform better than the self-supervised features, and the combination of all three yields the best results on both transfer problems and metrics.
In particular,  when comparing random initialization and the combined class affinities for  OASIS,  we  improve FID from 54.0 to 40.9 and mIoU from 30.0 to 31.4 when transferring from COCO to ADE, and  improve FID from 82.9 to 53.7 and  mIoU from 15.9 to 17.4 in the reverse tranfer direction.  
For PITI, we improve FID from 57.1 to 40.7 and mIoU from 11.6 to 22.3 when transferring from COCO to ADE, and improve  FID from 83.7 to 46.8 and mIoU from 0.8 to 7.5 in the opposite direction. 

While the FID values for the diffusion-based PITI model are comparable or better than those obtained using OASIS, we noticed that mIoU values for the diffusion-based model are worse. 
This trend was already observed on the source dataset:  PITI trained on the full COCO dataset  has an mIoU of 34.4, while the mIoU on OASIS is 44.1. This gap widens when training on the full ADE dataset, where PITI has an mIoU of 26 compared to 48.8 for  OASIS.

\mypar{Qualitative results} In Figure \ref{fig:ade20krndvssim}, we show samples synthesized from PITI and OASIS models with three different types of transfer, from ADE to COCO on top left, from COCO to ADE on top right and from COCO to Cityscapes on the bottom, finetuning with 100 target images with and without class affinity transfer (\ours). Both for PITI and OASIS, training with \ours leads to synthesized images with sharper details  and  better recognizable objects. For instance, the sink and bathtub in images of the first row in the top right of the figure are of better quality and more realistic when trained with \ours. Furthermore, when transferring to the challenging COCO dataset containing 183 classes, we notice that without \ours, PITI fails to synthesize realistic images adhering to the label maps, whereas \ours can synthesize images of better quality coherent with label maps.

\mypar{Comparison with the state of the art}
 We compare to the state-of-the-art transfer methods for generative models. 
 We consider four pairs of target-source datasets, by taking source models trained either on COCO-Stuff or ADE20K datasets, and finetuning them on target datasets of 100 images taken from  Cityscapes, ADE20K and COCO-Stuff. 

 From the results in  \tab{comp2baselines} using the OASIS architecture,  we observe a significant improvements with \ours. 
 We improve FID from 64.9  to 40.9 and mIoU from 26.2 to 31.4  COCO\textrightarrow ADE, and improve FID from 89.8   to 53.7 and mIoU from 15.0 to 17.4 for  ADE\textrightarrow COCO \wrt the best baseline cGAN-Transfer. 
 When transferring to Cityscapes, 
\ours  
improves FID from 56.2 (TransferGAN) to 51.4 and  mIoU from 62.7 (FreezeD) to 66.1 when the source dataset is ADE, and improving  FID from 49.8  to 47.0  and  mIoU from 66.5 to 68.1 \wrt FreezeD with COCO as source.
 
 In the case of PITI, directly finetuning  a pretrained GLIDE model on the target dataset (``From GLIDE'') produces  images with much better quality compared to OASIS trained from scratch in terms of FID. 
 However, the model fails to generate images with strong adherence to label maps, as reflected by the poor mIoU scores. 
 Generally, finetuning PITI trained on the source dataset on the target dataset (``finetune all'') gives better results than finetuning from GLIDE. 
\ours significantly improves  over these two baselines in all settings: with more than 15 points in FID, and more than 3 points in mIoU.  

\begin{figure}
\def\myim#1{ \includegraphics[width=27mm,height=23mm]{figure/diffSizeplots/#1}}
\centering
\setlength\tabcolsep{1 pt}
\begin{tabular}{ccc}
\myim{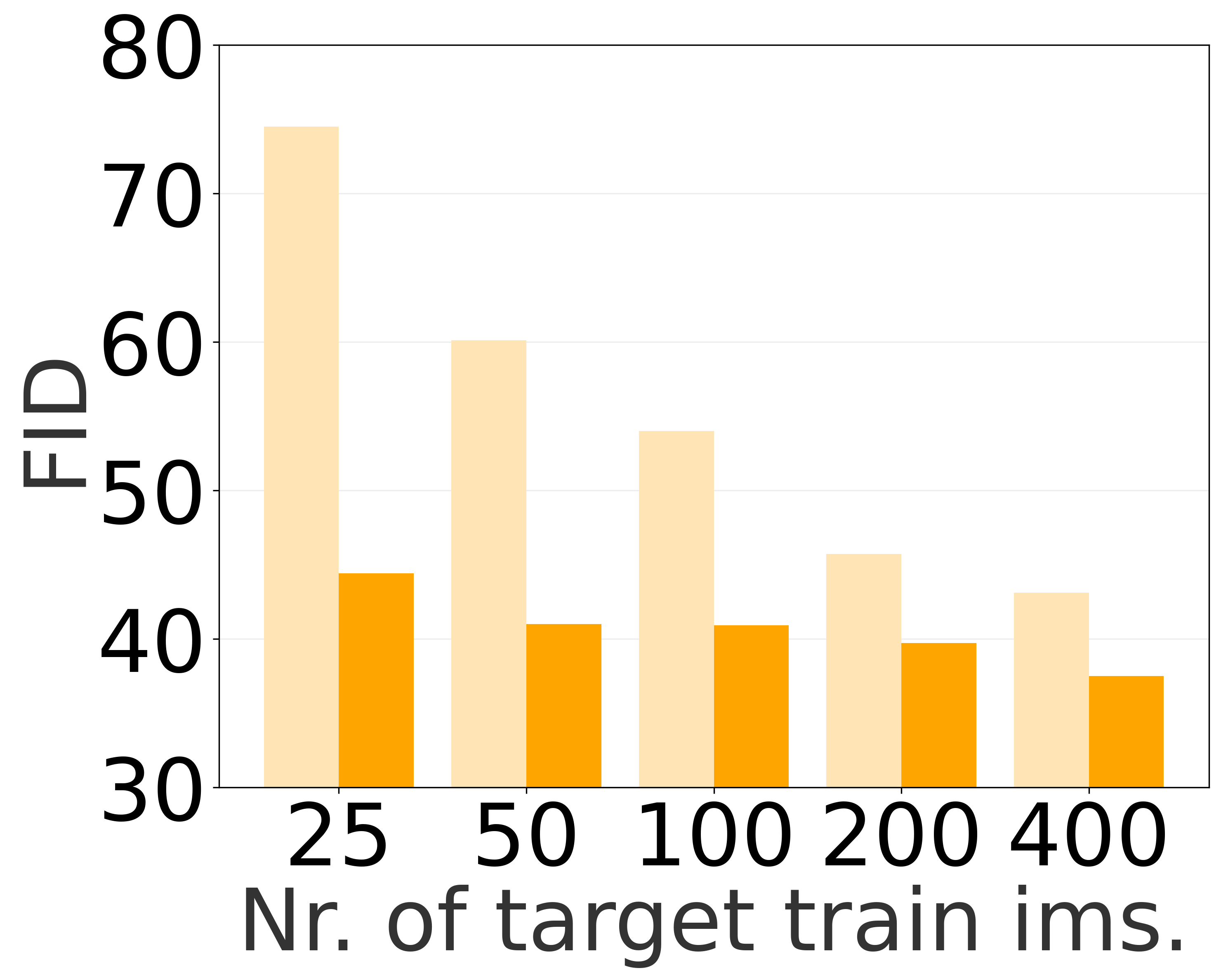} &
\myim{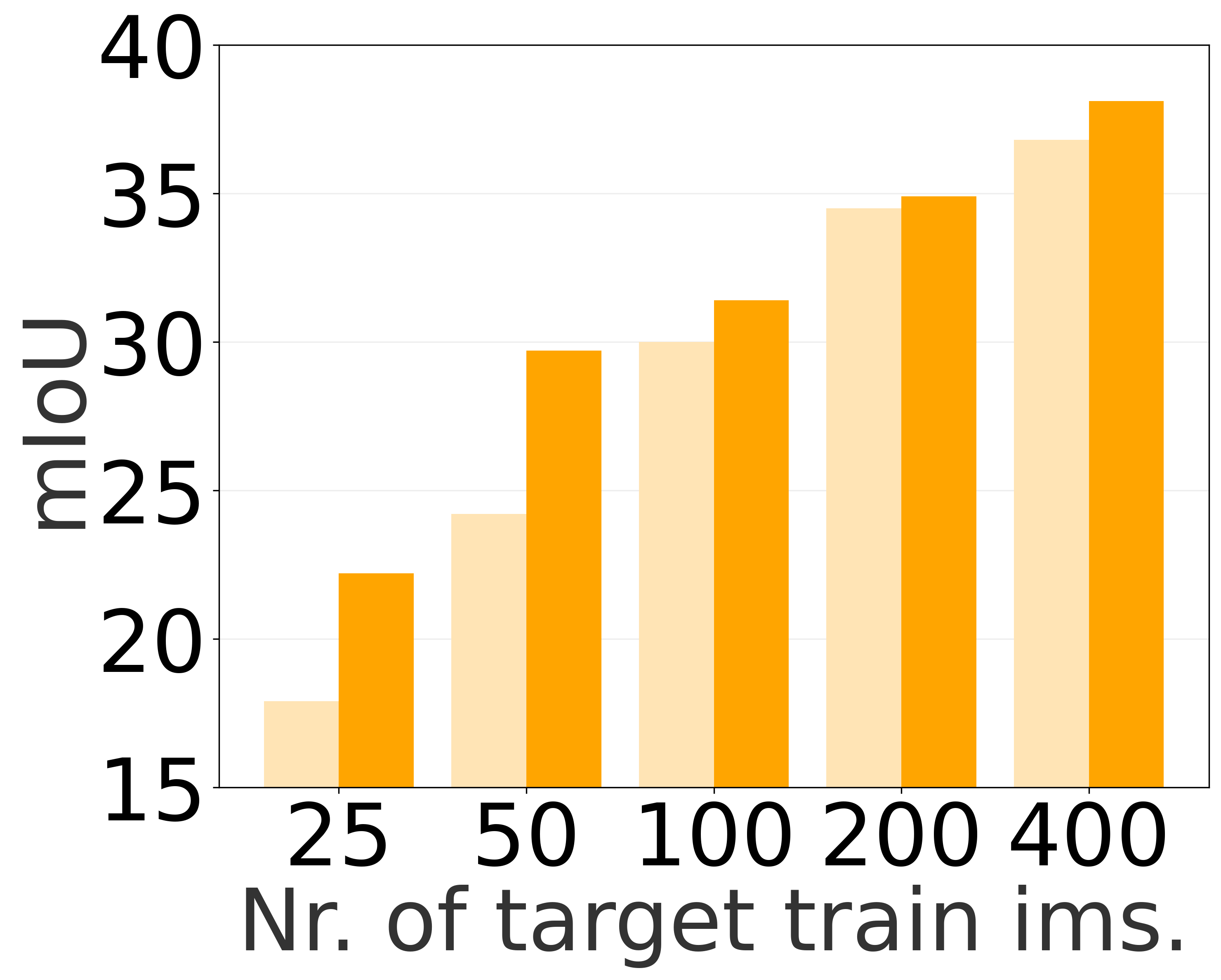} & 
\myim{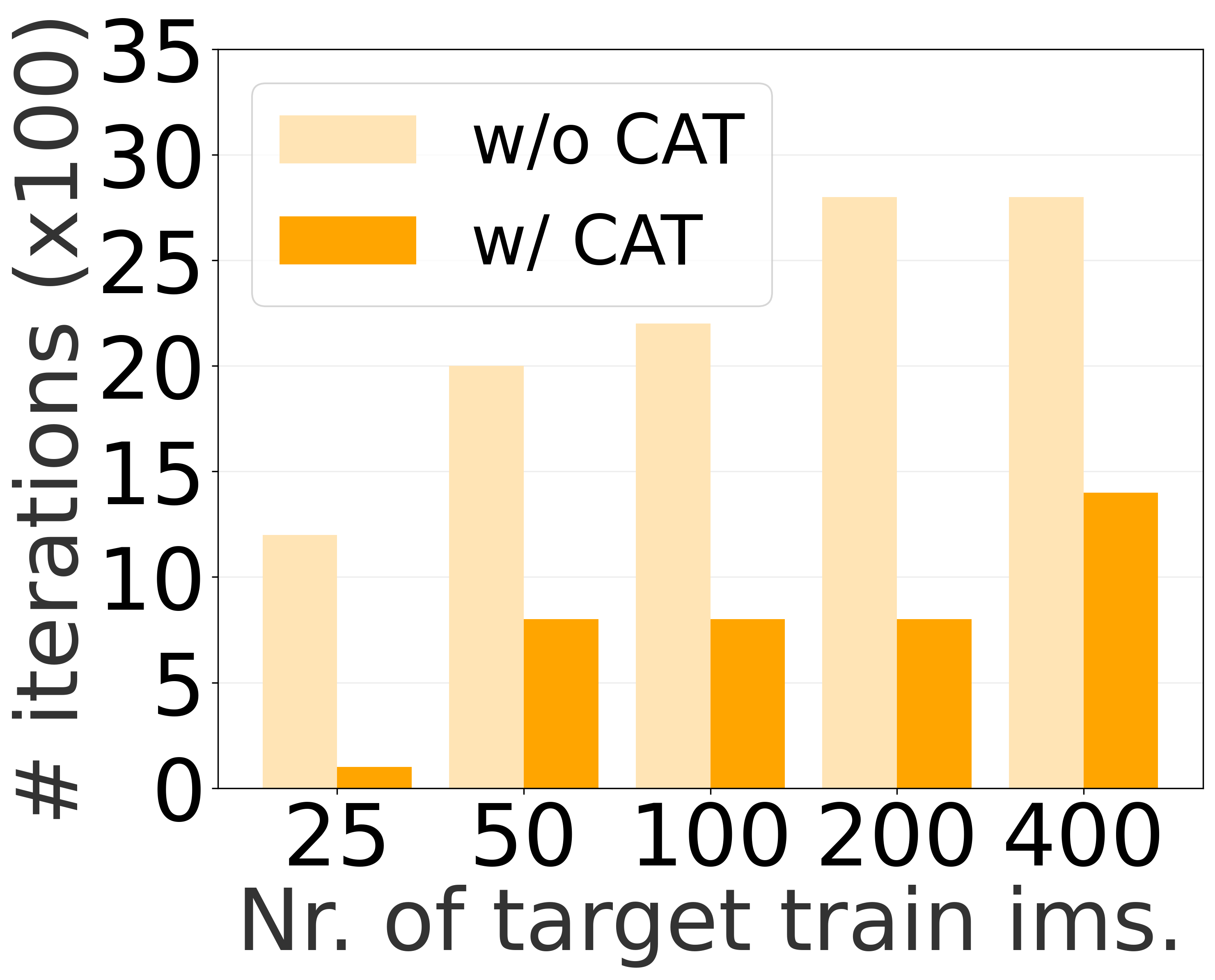} 
\end{tabular}
\myvspace
\caption{FID, mIoU and \# training iterations for COCO\textrightarrow~ADE transfer using OASIS w/ and w/o \ours for different dataset sizes.
}
\vspace{-0.5cm}
\label{fig:plotDifSize}
\end{figure}

\subsection{Ablation study and analysis}
\label{sec:setsize}

\mypar{Impact of the target dataset sizes}
We generate subsets of size ranging from 25 to 400 images on ADE and analyze the peformance of \ours using OASIS as base model. We report the evolution of FID and mIoU in Figure \ref{fig:plotDifSize} (left and center panel, respectively). While \ours (in dark orange) demonstrates gains in both FID and mIoU in all the dataset sizes, its advantage is striking in the very low-shot setting. \ours surpasses by large margins of 30.1 and 19.1 FID points, respectively on datasets of size 25 and 50. Even if the boost narrows when increasing the dataset size, we still observe a performance gain of 5.6 points of FID with 400 training images.
For reference, training OASIS on the full dataset ADE (with face blurring) achieves 29.8 of FID and 48.6 of mIoU. 
In the right panel of \fig{plotDifSize} we  report the number of training iterations before convergence. 
We note a faster convergence with \ours  for all the dataset sizes.

\input{tab3.tex}

\mypar{Training-free transfer} 
If we only add the class affinity matrix to the source model, without further finetuning the model on the target data, it can already be used to generate samples for the target domain. 
Since the class-affinity matrix is obtained with a single feed-forward pass through the target training data, or even just using the textual class embeddings, this approach can be considered ``training-free'' and is extremely computationally efficient. 

From the results in \tab{0shotgen}, we notice that the affinity matrix that combines the different methods consistently obtains the best (or very close)  performance in terms of FID and mIoU. 
No matter the choice of affinity estimation method, our training-free variant of \ours achieves performance far better than using a randomly initialized class affinity matrix.
Both in terms of FID and mIoU, the training-free result is already relatively close  to the results obtained with finetuning (``Combo + finetuning'').
This underlines the key role of a good initialization for transfer for semantic image synthesis. 
Qualitative results of training-free transfer can be found in the supplementary material.

\mypar{Ablations for OASIS}
In Table \ref{tab:ablationoasis}, we ablate how each component contributes to the performance. We observe that freezing part of the discriminator parameters, as done in~\cite{mo20cvpr}, improves performance. We also found it beneficial to perform finetuning in two stages: in the first stage, we fix most of our generator parameters, and only finetune the first convolution in each SPADE block that takes as input the segmentation map; in the second stage, we train all the generator parameters. We also demonstrate that adding residual convolutional layers is beneficial (\eg~ from 83.4 to 79.9 in FID in ADE\textrightarrow COCO). 
Finally, we obtain substantial gains using our class affinity matrix for initialization.

\mypar{Ablations for PITI}
The ablation study in the case of the diffusion-based model is reported in Table~\ref{tab:ablationpiti}. We observe better performance when fixing the decoder part, and introducing  trainable prompts in the segmentation map encoder further improves FID and mIoU. Lastly, when adding our class affinity matrix, we consistently improve performance by a larger margin in all settings according to both metrics.

\begin{table}
\centering
\setlength{\tabcolsep}{2.5pt} 
{
\small
\begin{tabular}{ccccccccc}
\toprule
   \multirow{2}{*}{\textbf{FreezeD}} & \multirow{2}{*}{\textbf{2 stages}}  & \multirow{2}{*}{\textbf{Resid.}}& \multirow{2}{*}{\textbf{CAT}} &  \multicolumn{2}{c}{\textbf{COCO\textrightarrow ADE  }} &  \multicolumn{2}{c}{\textbf{ADE\textrightarrow COCO  }}
  \\
  &  &  &  & 
 $\downarrow$\textbf{FID}  &    $\uparrow$\textbf{mIoU}  & 
 $\downarrow$\textbf{FID}   &    $\uparrow$\textbf{mIoU}    \\  

\midrule
\xmark & \xmark & \xmark & \xmark  &87.2 & 20.7 &  117.3 & 11.0\\ 
\cmark & \xmark & \xmark & \xmark  &65.7 & 25.8 & 98.6 &14.8\\ 

\cmark & \cmark & \xmark & \xmark &  55.9 & 28.6 &  83.4 &15.2 \\ 
\cmark & \cmark & \cmark &  \xmark & 55.2 & 29.4 &79.9 & 15.7 \\
\cmark & \cmark & \cmark & \cmark  &  \textbf{40.9} &  \textbf{31.4}   &  \textbf{53.7}  & \textbf{17.4}  \\

\bottomrule

\end{tabular}
}
\caption{Ablations with adversarial OASIS architecture.
}

\label{tab:ablationoasis}
\end{table}

\begin{table}

\centering
\setlength{\tabcolsep}{5pt} 
{
\small
\begin{tabular}{ccccccccc}
\toprule
   \multirow{2}{*}{\textbf{FixDec}}   & \multirow{2}{*}{\textbf{Prompts}} &   \multirow{2}{*}{\textbf{CAT}} &  \multicolumn{2}{c}{\textbf{COCO\textrightarrow ADE  }}&  \multicolumn{2}{c}{\textbf{ADE\textrightarrow COCO  }}
  \\
  &&   & 
 $\downarrow$\textbf{FID}  &    $\uparrow$\textbf{mIoU}  &
 $\downarrow$\textbf{FID}   &    $\uparrow$\textbf{mIoU}   \\  

\midrule
  \xmark & \xmark & \xmark  & 56.5 &13.5& 85.0 & 0.1\\ 
  \cmark & \xmark & \xmark  & 52.4 & 13.6 &  79.0 & 1.4\\ 
\cmark & \cmark & \xmark  & 51.1 & 14.1 &   78.9 & 1.3\\ 

\cmark & \cmark & \cmark  & \bf 40.7 & \bf 22.3 &  \bf 46.8 & \bf  7.5 \\
\bottomrule
\end{tabular}
}
\caption{Ablations with diffusion-based PITI architecture.
}
\vspace{-0.5cm}
\label{tab:ablationpiti}
\end{table}

%% file: fig3.tex
\begin{figure*}
 \def\myim#1{ \includegraphics[width=17.0mm,height=17.0mm]{samples/#1}}
 \def\myimCS#1{ \includegraphics[width=34.5mm,height=17.25mm]{samples/#1}}
     \centering
   \setlength\tabcolsep{0.5 pt}
   \renewcommand{\arraystretch}{0.2}
     \begin{tabular}{cccccccccccc}
&   \multicolumn{2}{c}{OASIS}&   \multicolumn{2}{c}{PITI}&\hspace{0.1cm}&&   \multicolumn{2}{c}{OASIS}&   \multicolumn{2}{c}{PITI}\\
\\
Input & w/o CAT  & w/ CAT & w/o CAT  & w/ CAT &\hspace{0.1cm}&Input & w/o CAT  & w/ CAT & w/o CAT  & w/ CAT\\
\myim{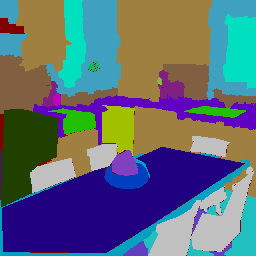} &
\myim{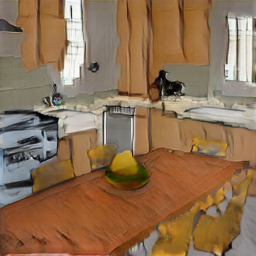} &
\myim{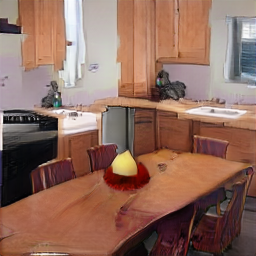} & 
\myim{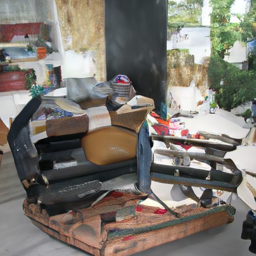} &
\myim{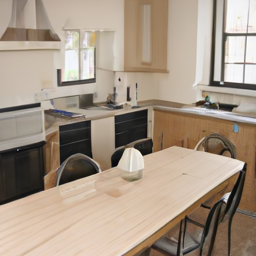} 
&\hspace{0.1cm}&
\myim{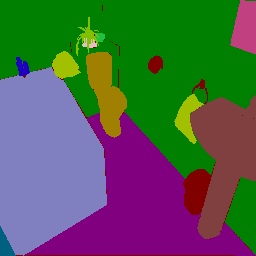} &
\myim{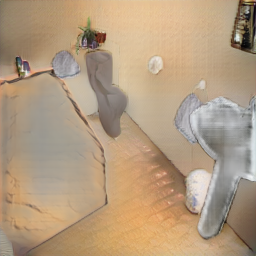} &
\myim{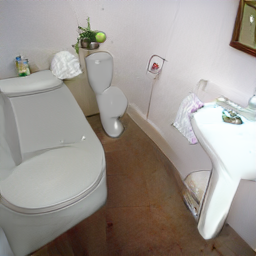} &
\myim{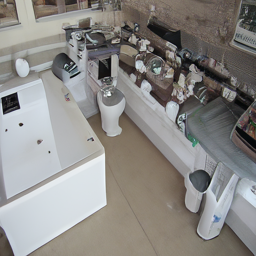} &
\myim{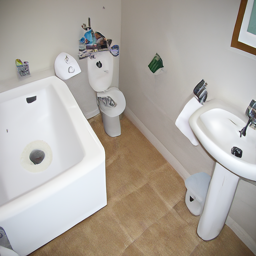} \\
\myim{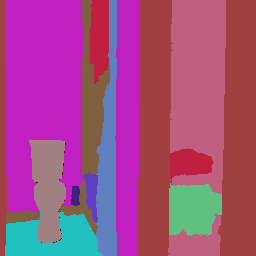} &
\myim{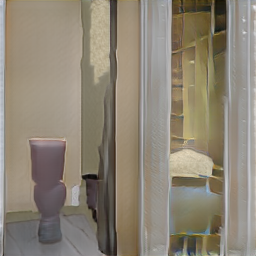} &
\myim{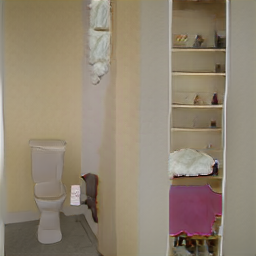} &
\myim{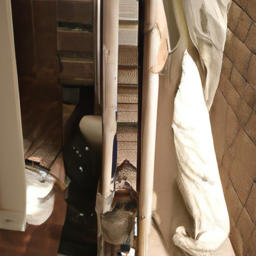} &
\myim{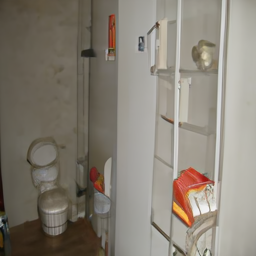}
&\hspace{0.1cm} &
\myim{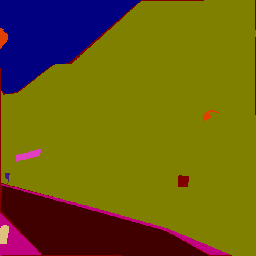} &
\myim{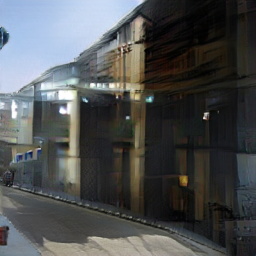} &
\myim{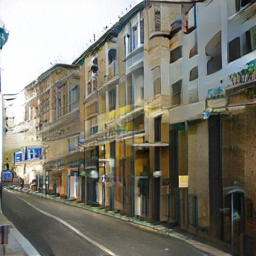} &
\myim{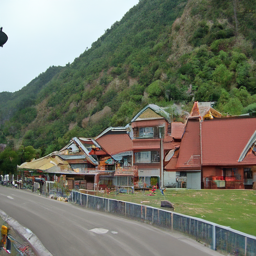} &
\myim{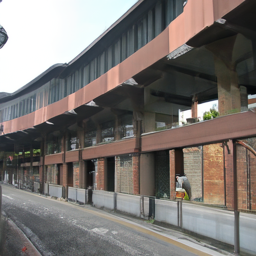} \\
\myim{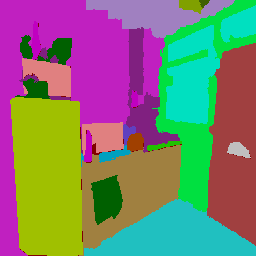}&
\myim{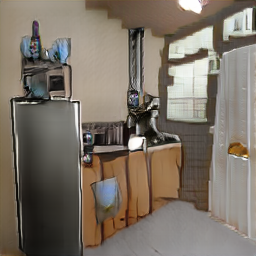}&
\myim{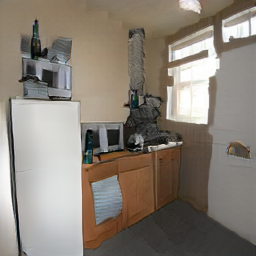}&
\myim{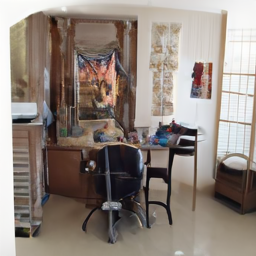}&
\myim{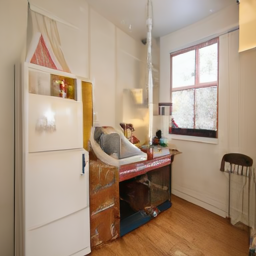}
&\hspace{0.1cm}&
\myim{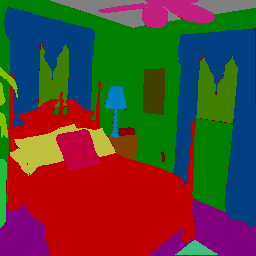} &
\myim{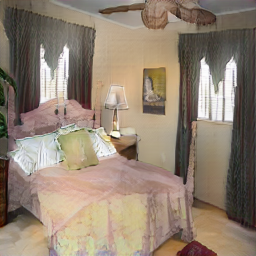} &
\myim{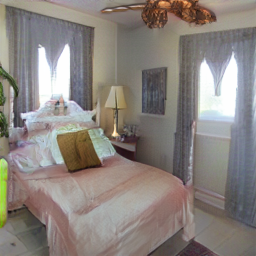} &
\myim{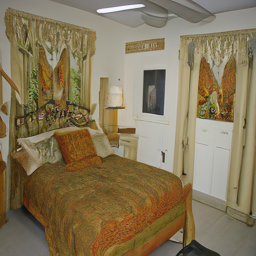} &
\myim{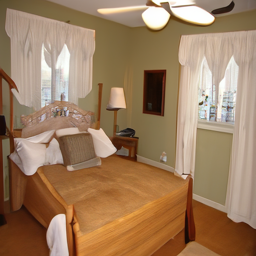}    \\
\end{tabular}
\\
\vspace{5mm}

\begin{tabular}{ccccc}
&   \multicolumn{2}{c}{OASIS}&   \multicolumn{2}{c}{PITI}\\
\\
Input & w/o CAT  & w/ CAT & w/o CAT  & w/ CAT \\
\myimCS{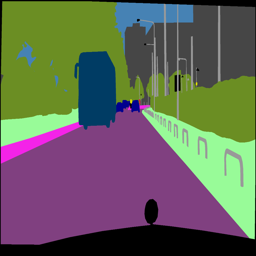} &
\myimCS{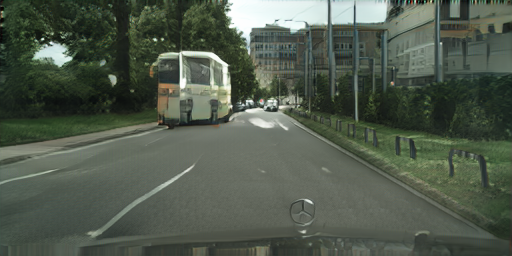} & 
\myimCS{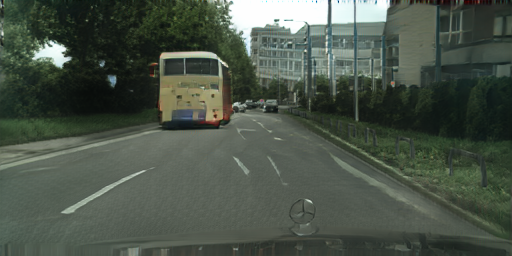}&
\myimCS{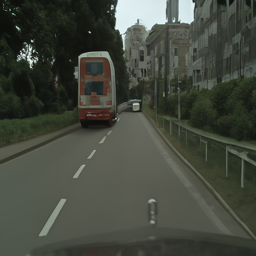} & 
\myimCS{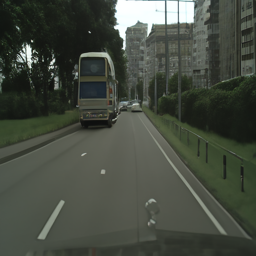}\\
\myimCS{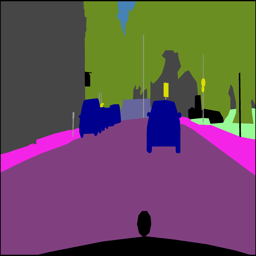}&
\myimCS{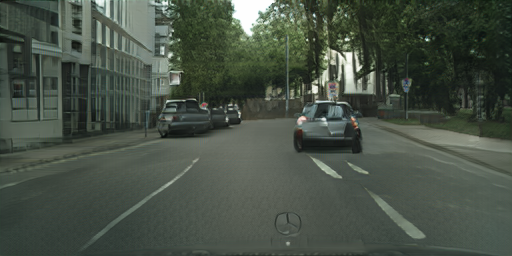} &
\myimCS{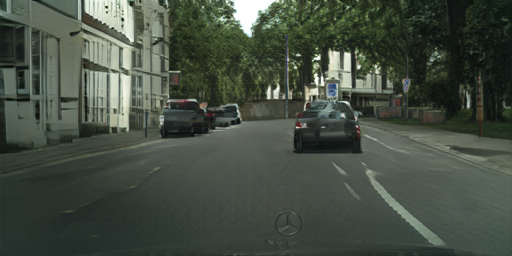}&
\myimCS{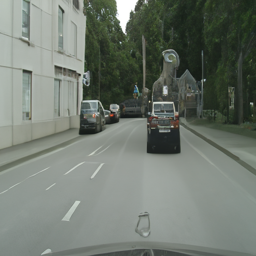} &
\myimCS{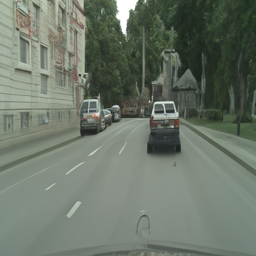}\\
\myimCS{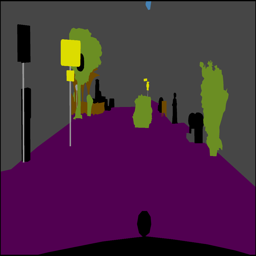}&
\myimCS{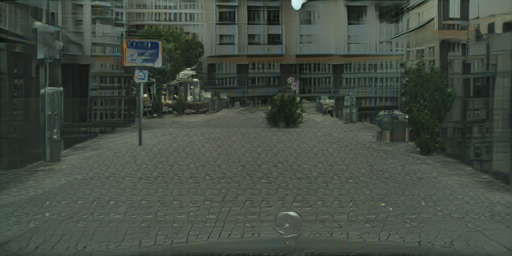} &
\myimCS{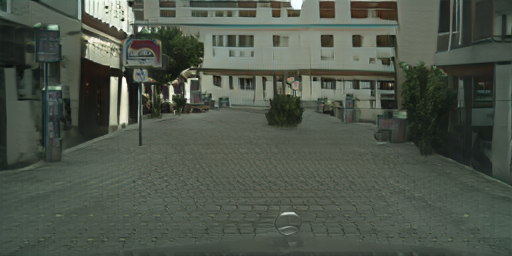}&
\myimCS{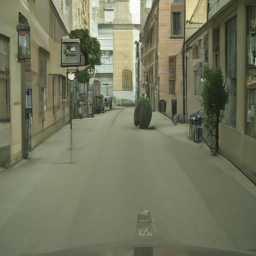} &
\myimCS{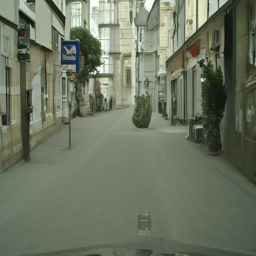}
\end{tabular}
\myvspace
\caption{Samples from models trained with 100 target images.
Transfer from ADE to COCO (top left), COCO to ADE (top right), and from COCO to Cityscapes (bottom).
Class affinity matrix initialized randomly (w/o CAT) or with   combination method (w/ CAT). 
}
\label{fig:ade20krndvssim}
\end{figure*}

%% file: table1tall.tex
\begin{table}
\centering
 \setlength{\tabcolsep}{5pt} 
\small
{
\begin{tabular}{llcccc}
\toprule
  & \textbf{Affinity matrix}&  \multicolumn{2}{c}{\textbf{COCO\textrightarrow ADE  }}
  &  \multicolumn{2}{c}{\textbf{ADE\textrightarrow COCO  }}  \\
 &   \textbf{initialization}&
 $\downarrow$\textbf{FID~}  &    $\uparrow$\textbf{mIoU}  &
 $\downarrow$\textbf{FID~}   &    $\uparrow$\textbf{mIoU}  \\  
\midrule

\multirow{ 5}{*}{\begin{sideways}\textbf{OASIS}\end{sideways}} & Random  & 54.0 & 30.0 &  82.9 & 15.9 \\

  & Text-based    &  41.1 &  30.4 &  55.2 &   17.3\\
  
  & Segmentation   &  42.0 &  30.8 &  58.2 &  12.9 \\
  
    & Self-supervised  &  41.3 &  29.8 &  57.9 &  15.4\\
   
  & Combination  & \textbf{40.9} &  \textbf{31.4} &  \textbf{53.7}  & \textbf{17.4}\\
  \midrule
  \multirow{ 5}{*}{\begin{sideways}\textbf{PITI}\end{sideways}}  & Random &  57.1 &11.6 & 83.7 & 0.8\\ 

    & Text-based  & 40.9 &  20.2& 47.4 & 7.1 \\
 
   & Segmentation  & 41.1 & 22.0 & 52.5 &  5.3\\

   & Self-supervised & 41.9 & 21.2 & 50.9 & 5.6 \\
  
   & Combination    &  \bf 40.7 &   \bf 22.3 & \bf   46.8 &  \bf  7.5\\
\bottomrule
\end{tabular}
}
\caption{Comparison of different class affinity estimation methods and their combination to randomly initializing the affinity matrix. Results after finetuning.
}
\label{tab:effectClsSimExtrc}
\end{table}

%% file: tab2.tex
\begin{table*}
\centering
\small
 \setlength{\tabcolsep}{5pt} 

{
\begin{tabular}{llcccccccc}
\toprule
    & \multirow{ 2}{*}{\textbf{Method}} &  \multicolumn{2}{c}{\textbf{COCO\textrightarrow ADE  }}
  &  \multicolumn{2}{c}{\textbf{ADE\textrightarrow COCO  }}
   &  \multicolumn{2}{c}{\textbf{ADE\textrightarrow Cityscapes  }} & \multicolumn{2}{c}{\textbf{COCO\textrightarrow Cityscapes  }}
  \\ 
   &&

 $\downarrow$\textbf{FID~}  &    $\uparrow$\textbf{mIoU}  &
 $\downarrow$\textbf{FID~}   &    $\uparrow$\textbf{mIoU} &
 $\downarrow$\textbf{FID}  &    $\uparrow$\textbf{mIoU} &
 $\downarrow$\textbf{FID~}  &    $\uparrow$\textbf{mIoU}   \\  
\midrule

\multirow{7}{*}{\begin{sideways}\textbf{OASIS}\end{sideways}}& From scratch & 145.9 & 13.6 & 153.4 & 7.1 &136.5 & 37.1 & 137.0 &37.2\\ 

& TransferGAN~\cite{wang18eccv2} & 85.1 & 20.4 & 120.5 & 10.2 &  56.2  &  61.5&  51.5 &  63.6 \\ 
& FreezeD~\cite{mo20cvpr}  & 66.3 & 25.9 & 102.4 & 13.8 &57.1 & 62.7&  49.8 & 66.5\\ 
& MineGAN~\cite{wang20cvpr}  &82.2& 21.0 & 110.2 & 11.5 & 57.8 & 62.2 &52.6 &65.4 \\ 
& BSA~\cite{noguchi19iccv}  & 70.1 & 25.9 & 94.2 & 12.8 &76.3 & 51.6 & 65.7 & 59.1\\

& cGAN-Transfer~\cite{shahbazi21cvpr}  & 64.9&  26.2 & 89.8& 15.0 &63.6& 61.7& 57.3 & 58.9 \\ 

& \ours (ours)   & \textbf{40.9} &  \textbf{31.4} &  \textbf{53.7}  & \textbf{17.4} & \bf 51.4 &  \textbf{66.1} &  \textbf{47.0} &  \textbf{68.1} \\

\midrule

\multirow{ 3}{*}{\begin{sideways}\textbf{PITI}\end{sideways}}  & From GLIDE~\cite{wang2022semantic} & 59.8& 2.0 & 104.9 &  0.3 &74.5 & 9.5& 74.5 &  9.5\\
 & Finetune all & 56.8 & 14.2&83.7 &  0.1 &86.1 & 17.2& 70.8 & 36.7\\

 & \ours (ours)  & \bf 40.7 & \bf 22.3 & \bf 46.8 & \bf  7.5& \bf 62.7 & \bf 27.3  & \bf 54.7  & \bf 39.9\\
\bottomrule
\end{tabular}
}
\caption{Comparison with state-of-the-art transfer methods, using  target datasets of 100 images. }

\vspace{-0.3cm}
\label{tab:comp2baselines}
\end{table*}

%% file: tab3.tex
\begin{table}
\centering
\setlength{\tabcolsep}{5pt} 
{\small
\begin{tabular}{llcccc}
\toprule
  &  \textbf{Affinity matrix } &  \multicolumn{2}{c}{\textbf{COCO\textrightarrow ADE  }} &  \multicolumn{2}{c}{\textbf{ADE\textrightarrow COCO  }}
  \\
  &  \textbf{initialization} &
 $\downarrow$\textbf{FID}  &    $\uparrow$\textbf{mIoU}  &
 $\downarrow$\textbf{FID}   &    $\uparrow$\textbf{mIoU}   \\  
\midrule 
 \multirow{ 6}{*}{\begin{sideways}\textbf{OASIS}\end{sideways}}
 & Random  & 216.0 & 0.5 & 270.8  & 0.1\\
 & Text-based     & 44.6  & 23.9  & 57.8  &  15.2\\
  & Segmentation   & 47.0  & 22.8  & 64.5  & 12.1 \\
  & Self-supervised  &45.5  & 22.7 & 68.1 & 10.2    \\
  
& Combination   & \textbf{43.1}  & \textbf{25.1} & \textbf{56.3}   & \textbf{13.8}  \\

& \emph{Combo + finetuning}  & \emph{40.9} &  \emph{31.4} &  \emph{53.7}  & \emph{17.4}  \\   
\midrule
 \multirow{ 6}{*}{\begin{sideways}\textbf{PITI}\end{sideways}} 
 & Random &  96.3 &  5.7 &  98.3&  0.1 \\ 
 & Text-based  & 51.6  & 19.2 & 50.8 &7.1\\
  & Segmentation  &  48.7& 20.2 & 59.2 &5.2\\
  & Self-supervised  & 49.5 & 19.6  & 53.9 &5.0\\
  & Combination   &  \bf 48.5 &  \bf 20.9 &   \bf 49.3 &   \bf 7.4 \\
  
  & \emph{Combo + finetuning}   &   \emph{40.7} &   \emph{22.3} &   \emph{46.8} &    \emph{7.5} \\
\bottomrule
\end{tabular}
}
\caption{Training-free transfer results. Results obtained with additional finetuning are marked in italic.
}
\vspace{-0.5cm}
\label{tab:0shotgen}
\end{table}

%% file: conc.tex
\section{Conclusion}
In this paper, we consider the problem of few-shot  semantic image synthesis, where training sets consist of a few tens to a few hundreds images.
To address this problem, we proposed \textbf{C}lass \textbf{A}ffinity \textbf{T}ransfer (\ours), a transfer learning approach
 based on estimating a class affinity matrix, using the similarities among classes in the source and target datasets.
We consider four methods to establish these similarities: based on a semantic segmentation model of the source domain, using self-supervised vision features, or using  text-based class label embeddings, and a combination via majority voting.
The class affinity matrix is prepended as a first layer to the source model to align it with the one-hot-labels of the target domain.
We integrated our approach in both an adversarial (OASIS) and a diffusion-based architecture (PITI). We conducted extensive experiments on the COCO-Stuff, ADE20K, and Cityscapes datasets, and observed excellent transfer performance. Consistently outperforming state-of-the-art transfer methods for generative models, and allowing realistic semantic image synthesis using training sets as small as 100 images.

%% file: supmat.tex
\appendix

In this supplementary material, we  provide additional details regarding the  code and datasets used in our experiments in Section~\ref{app:assets}. 
In Section~\ref{app:archi}, we detail the proposed used architectures. 
In Section~\ref{app:implem}, we provide additional implementation details. 
In Section~\ref{sec:app_affinities} we list the class correspondences between source and target found by our method.
We  report extended quantitative results in Section~\ref{app:quantitative}, and results, and provide  additional qualitative results in Section~\ref{app:qualitative}.

\section{Assets and licensing information}
\label{app:assets}
In Table~\ref{table:assets}, we provide the links to the  datasets and code repositories we used, and list their licenses in Table~\ref{table:licenses}.

To avoid training our generative models on sensitive personal data, we train our models on filtered data. 
For Cityscapes, we use the version of the dataset in which human faces and license plates have been blurred, as provided through the dataset distribution website.
For ADE20K and COCO, we processed the data ourselves to detect human faces, segment them, and blur them. 
We find that overall this processing has limited impact on performance. 
For the OASIS source models trained on COCO and ADE20K: we find that the FID changes from 17.0 to 18.7 for COCO, and from 28.3 to 29.8 for ADE20K. 
The FID metric is still computed \wrt the original non-filtered datasets.

\begin{figure}
\centering
\includegraphics[width=\columnwidth]{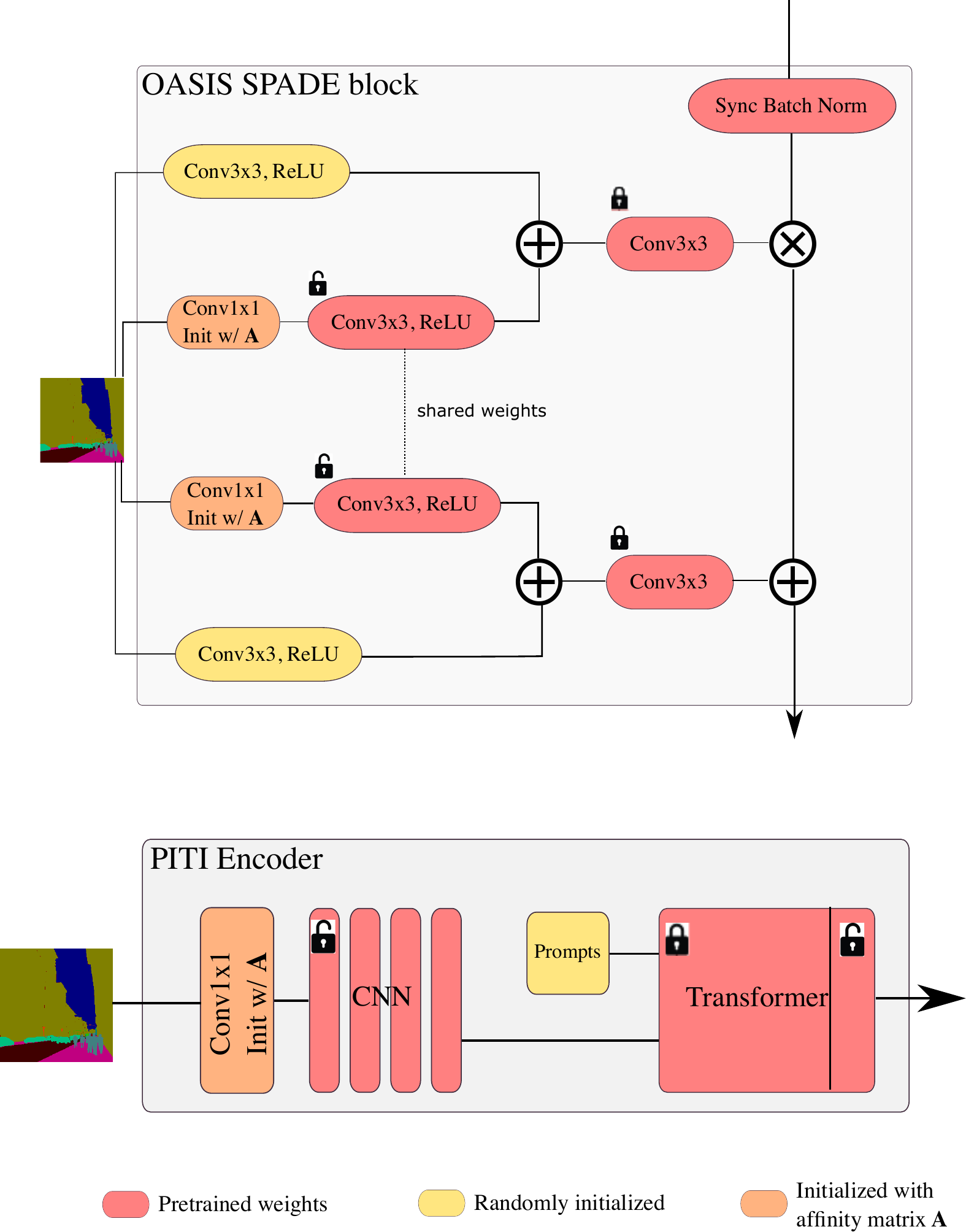}
\caption{
Overview of our modifications of the OASIS and PITI architectures for semantic image synthesis.
The class affinity matrix is used to align the source model with the target label space, and then models are further finetuned using the  target data.
}
\label{fig:drawingOASISPITI}
\end{figure}

\begin{table*}
\centering
\scriptsize
\begin{tabular}{cc}
\toprule
\textbf{Dataset }\\
COCO-Stuff~\cite{caesar18cvpr} & \url{https://cocodataset.org/} \\ 
Cityscapes~\cite{cordts16cvpr} & \url{https://www.cityscapes-dataset.com/} \\ 
Ade20K~\cite{zhou17cvpr}& \url{http://groups.csail.mit.edu/vision/datasets/ADE20K/}\\
\midrule
\textbf{Model}\\
OASIS~\cite{sushko21iclr} &  \url{ https://github.com/boschresearch/OASIS} \\
PITI~\cite{wang22arxiv} &  \url{https://github.com/PITI-Synthesis/PITI} \\
\bottomrule
\end{tabular}
\caption{Links to the assets used in the paper.}
\label{table:assets} 
\end{table*}

\begin{table*}
\centering
\scriptsize
\begin{tabular}{cc}
\toprule
\bf Dataset\\ 
COCO-Stuff~\cite{caesar18cvpr} & \url{https://www.flickr.com/creativecommons} \\ 
Cityscapes~\cite{cordts16cvpr} & \url{https://www.cityscapes-dataset.com/license} \\ 
Ade20K~\cite{zhou17cvpr}& MIT License\\
\midrule
\textbf{Model} \\
OASIS~\cite{sushko21iclr} &  \url{https://github.com/boschresearch/OASIS/blob/master/LICENSE} \\
PITI~\cite{wang22arxiv} &  \url{https://github.com/PITI-Synthesis/PITI/blob/main/LICENSE} \\
\bottomrule
\end{tabular}
\caption{Assets licensing information.}
\label{table:licenses}
\end{table*}

\section{Architectural designs}
To complement the description of our method provided in the main paper, we  report additional technical details.
\label{app:archi}

\mypar{Architectural design to adversarial model}
In Figure~\ref{fig:drawingOASISPITI} (top panel), we illustrate the modifications to SPADE blocks used in the OASIS generator to include our class affinity matrix. 
In red, we display original blocks from OASIS generator which are initialized with pretrained weights from the source model during finetuning on a target dataset. We prepend a $1\!\times\!1$ convolution indicated in orange blocks initialized with our affinity class matrix $\Amat$. The residual layers are shown in yellow and are initialized with zeros. They take as input the target segmentation maps. 
During the first finetuning stage, only red blocks with open padlock in the drawing as well as all the orange and yellow blocks are finetuned, while in a second stage we train all layers in the spade blocks.  

We provide an ablation in Table~\ref{tab:linlayerAblationoasis} on whether using shared or separate paths to  compute scale and shift parameters. 
When shared, the modules in the SPADE block are not duplicated, and the last  $3\!\times\!3$ convolutional layer outputs both the scale and shift parameters.
When separate, all blocks are duplicated as in Figure~\ref{fig:drawingOASISPITI} (top panel), but weights are shared on the first $3\!\times\!3$  convolutional layer.
The results indicate benefit in mIoU and FID  when using separate paths, and we use this option in our main experiments.

\begin{table}
\centering
 \setlength{\tabcolsep}{5pt} 
{\small
\begin{tabular}{ccc}
\toprule
\multirow{2}{*}{\textbf{Paths}}  &  \multicolumn{2}{c}{\textbf{COCO\textrightarrow ADE  }} \\
&  $\downarrow$\textbf{FID}  &    $\uparrow$\textbf{mIoU}    \\  
 \midrule
Shared &41.4&30.1 \\
Separate & \bf 40.9 & \bf 31.4\\
\bottomrule
\end{tabular}
}
\caption{Ablation on using shared or separate paths in the SPADE blocks of the modified OASIS architecture.
}
\label{tab:linlayerAblationoasis}
\end{table}

\mypar{Architectural changes to diffusion model} In Figure~\ref{fig:drawingOASISPITI} (bottom panel), we display our changes to the PITI encoder in orange and yellow blocks. 
Similarly as for OASIS, we prepend  a $1\!\times\!1$ convolution initialized by the class affinity matrix. 
Then, as specified in the method section from the main paper, we add trainable ``prompt'' tokens randomly initialized  at the input of Transformer block. 
During training, we finetune these prompts, the $1\!\times\!1$ convolution, the CNN, as well as the last residual block of the Transformer.

\input{hardsoft}

\section{Complementary training details}
\label{app:implem}

When using the GAN-based OASIS architecture~\cite{sushko21iclr}, we  use the publicly available checkpoints of as our source models. For the diffusion-based PITI architecture~\cite{wang22arxiv},  we use the released checkpoint for COCO-Stuff. 
The checkpoint for ADE20K is not released, and we therefore trained the model ourselves using the public code, obtaining an FID of 27.8, comparable to the 27.3 reported by the authors. 

We monitor finetuning on target datasets by computing FID on the validation set and employ early stopping using this FID criterion.
We optimize learning rates on both OASIS and PITI. 
For PITI, we try learning rates $lr=\{3.5e-5, 7e-5, 1e-4, 2e-4\}$, and $ lr = \{1e-4,4e-4,8e-4\}$ for OASIS. 
For PITI, we choose a learning rate of $7e-5$ when transferring to Cityscapes, while we set $lr = 2e-4$ when doing transfer to ADE20K and COCO-Stuff. 
For OASIS, we use $lr = 4e-4$ when transferring to Cityscapes while we set $lr = 1e-4$ for ADE20K and COCO-Stuff. 

\section{Estimated class affinities}
\label{sec:app_affinities}

In Table~\ref{tab:classmappingCity2Cocox} and Table~\ref{tab:classmappingCity2Ade} we show the class mappings obtained for Cityscapes and ADE20K target classes with COCO-Stuff classes as source using the combination method. We see that most of the class correspondences are coherent, but there are a few aberrations. 
For instance, the ``truck'' class from Cityscapes is associated with ``building'' class from ADE20K. This could be explained by image style discrepancy existing between ADE20K and Cityscapes, as samples from Cityscapes contain darker images and less colorful than ADE20K, which is misleading image-based methods for class similarity estimation.

\begin{table}

\centering
 \setlength{\tabcolsep}{5pt} 
\resizebox{\columnwidth}{!}
{
\begin{tabular}{cc|cc}
\toprule
  \textbf{Cityscapes}&\textbf{COCO} &   \textbf{Cityscapes}&\textbf{COCO} \\
 
\midrule
unlabeled & sky-other & ego vehicle & road \\
rectification border & unlabeled & out of roi & unlabeled \\
static & building-other & dynamic & building-other\\
guard rail & railing& bus & bus\\
ground & pavement  & bridge &  building\\

road & road & pole & building-other \\
sidewalk & pavement  & polegroup & fence\\
parking & road & traffic light & traffic light \\
rail track & road  & traffic sign & builidng-other\\
building & building-other & vegetation & tree\\
wall & building-other  & terrain  & grass \\
fence & fence  & sky & sky-other \\
person & person  & rider & person \\
car & car  & truck & truck \\

caravan & truck  & trailer & truck \\
train & bus  & motorcycle & motorcycle \\
bicycle & bicycle  & license plate & unlabeled \\
tunnel & building\\

\bottomrule
\end{tabular}
}
\caption{Affinity class mappings for Cityscapes target classes with COCO-Stuff source classes using combination method.
\label{tab:classmappingCity2Cocox} 
}
\end{table}

\begin{table}

\centering
 \setlength{\tabcolsep}{5pt} 
 \resizebox{\columnwidth}{!}
{
\begin{tabular}{cc|cc}
\toprule

  \textbf{ADE}&\textbf{COCO} &   \textbf{ADE}&\textbf{COCO} \\
 
\midrule

wall & wall-concrete & building' & building-other \\
sky & sky-other & floor & floor-tile \\
tree &  tree & ceiling & ceiling-other \\
 road & road & bed & bed \\
windowpane & window-other & grass & grass \\
cabinet & cabinet & sidewalk & pavement \\
person & person & earth & dirt \\
door & door-stuff & table & table \\
mountain & mountain & plant & potted plant \\
curtain & curtain & chair & chair \\
car & car & water & water-other \\
painting & building-other & sofa & couch  \\
shelf & shelf & house & house' \\
sea & river & mirror & mirror-stuff \\
rug & rug & field& grass \\
armchair & couch & seat & chair \\
fence & fence & desk & desk \\
rock & rock & wardrobe & cabinet \\
lamp & light & bathtub & toilet \\
railing & clock & cushion & couch \\
base & house & box & plastic \\
signboard & street sign &chest & cabinet \\ 
counter & counter &sand & sand \\ 
sink & sink &skyscraper & skyscraper \\ 
fireplace & wall-stone &refrigerator & refrigerator \\ 
grandstand & platform &path & road \\ 
stairs & stairs &runway & road \\ 
case & counter &pool & boat \\ 
pillow & bed &screen & wall-tile \\ 
stairway & stairs &river & river \\ 
bridge & bridge &bookcase & shelf \\ 
blind & window-blind &coffee & table \\ 
toilet & toilet &flower & flower \\ 
book & book &hill & hill \\ 
bench & bench &countertop & counter \\ 
stove & oven &palm & tree \\ 
kitchen & dining table &computer & tv \\ 
swivel & chair &boat & boat \\ 
bar & desk &arcade & platform \\ 
hovel & wall-concrete &bus & bus \\ 
towel & towel &light & light \\ 
truck & truck &tower & building-other \\ 
chandelier & light &awning & tent \\ 
streetlight & building-other &booth & desk \\ 
television & tv &airplane & airplane \\ 
dirt & dirt &apparel & clothes \\ 
pole & metal &land & hill \\ 
bannister & railing &escalator & stairs \\ 
ottoman & furniture-other &bottle & bottle \\ 
buffet & counter &poster & mirror-stuff \\ 
stage & counter &van & car \\ 
ship & boat &fountain & sink \\ 
conveyer & oven &canopy & ceiling-other \\ 
washer & oven &plaything & teddy bear \\ 

\bottomrule
\end{tabular}
}
\caption{
Affinity class mappings for ADE20K target classes with COCO-Stuff source classes using combination method.
\label{tab:classmappingCity2Ade} 
}
\end{table}

\section{Additional quantitative results}
\label{app:quantitative}
In tables \ref{tab:effectClsSimExtrc2}, \ref{tab:0shotgen2}, \ref{tab:ablationoasis2}, \ref{tab:ablationpiti2}, we complement the quantitative results in the main paper by showing results when using Cityscapes as target dataset. 

We provide results  after finetuning  and in training-free mode in 
Table~\ref{tab:effectClsSimExtrc2} and Table~\ref{tab:0shotgen2} respectively. 
Similar to what was observed when transferring to ADE20K and COCO-Stuff, we see that in general the supervised and text-based methods perform better than the self-supervised one. Besides, the combination method obtains results that are among the best with respect to the different methods.

We  add ablations on PITI and OASIS in Table~\ref{tab:ablationoasis} and Table~\ref{tab:ablationpiti}.
The results are consistent, and show  the benefit of each component added in the OASIS and PITI pipelines.

\mypar{Hand-designed affinity matrix estimation} We did the ablation of manually assigning target classes to their most similar source classes in Table \ref{tab:handdesigned}. In COCO\textrightarrow ADE,  it requires manually checking $151\times 183$ pairs, while CAT automates this. With OASIS, we obtained performance similar to CAT: FID of 41.3 \vs 40.9 and a mIoU of 31.6 \vs 31.4.

\mypar{KID comparisons} In Table \ref{tab:KID}, we add KID metrics to further evaluate the benefit of finetuning compared to training-free mode on OASIS and PITI with ADE20K and COCO-Stuff dataset. We report improvements in all the settings after finetuning.

\begin{table}
\centering
\setlength{\tabcolsep}{5pt} 
{\small
\begin{tabular}{llc}
\toprule
    &  \multicolumn{2}{c}{\textbf{COCO \textrightarrow ADE}} 
  \\
  & $\downarrow$\textbf{FID}  &    $\uparrow$\textbf{mIoU}  \\
\midrule
 Computed  & \bf 40.9 & 31.4\\  

 Hand-designed & 41.3 & \bf 31.6\\  

\bottomrule
\end{tabular}
}
\caption{Ablation on the use of hand-designed affinity matrices \vs. ones computed with our Combination method. 
}
\label{tab:handdesigned}
\end{table}

\begin{table*}
\centering
 \setlength{\tabcolsep}{5pt} 
\resizebox{\textwidth}{!}
{\tiny
\begin{tabular}{lccccccccc}
\toprule
 & \textbf{Affinity matrix}&  \multicolumn{2}{c}{\textbf{COCO \textrightarrow ADE  }}
  &  \multicolumn{2}{c}{\textbf{ADE \textrightarrow COCO  }}
   &  \multicolumn{2}{c}{\textbf{ADE \textrightarrow Cityscapes  }}& \multicolumn{2}{c}{\textbf{COCO \textrightarrow Cityscapes  }}
  \\
 
 &   \textbf{initialization}&
 $\downarrow$\textbf{FID~}  &    $\uparrow$\textbf{mIoU}  &
 $\downarrow$\textbf{FID~}   &    $\uparrow$\textbf{mIoU} &
 $\downarrow$\textbf{FID~}  &    $\uparrow$\textbf{mIoU} &
 $\downarrow$\textbf{FID~}  &    $\uparrow$\textbf{mIoU}   \\  

\midrule
\multirow{6}{*}{\begin{sideways}\textbf{OASIS}\end{sideways}} & From scratch  & 55.0 & 29.8 &  79.8 & 17.8 & 55.3 & 64.0 & 47.6 & 66.0\\

  & Text-based    & 41.1 &  30.4 &  55.2 &   17.3 &  51.6 &  66.0& 47.7 &  67.3\\

   & Supervised   &  42.0 &  30.8 &  58.2 &  12.9  &51.8 &  65.7 & \bf 46.7 &  68.2\\

    & Self-supervised  &  41.3 &  29.8 &  57.9 &  15.4  & 53.0 &  65.1 &  47.5 &  67.6\\

  & Combination  & \textbf{40.9} &  \textbf{31.4} &  \textbf{53.7}  & \textbf{17.4} &  51.3 &  \textbf{66.4} &  47.0 &  \textbf{68.1}  \\

\midrule

 \multirow{5}{*}{\begin{sideways}\textbf{PITI}\end{sideways}}  & Random &  57.1 &  11.6 &  83.7 & 0.8 & 65.4 & 20.3&  58.5 & 33.1\\ 
 
  & Text-based  & 40.9 &  20.2& 47.4 & 7.1 & \bf 61.1 & \bf 28.5 &  57.1 & 38.0\\
  
   & Supervised  & 41.1 & 22.0 & 52.5 &  5.3 &63.0 & 26.7 & 57.3  & \bf 40.8\\
   & Self-supervised & 41.9 & 21.2 & 50.9 & 5.6 &63.8 &27.5 & 57.7  & 39.4\\
   & Combination    &   \bf 40.7 &   \bf 22.3 & \bf   46.8 &  \bf  7.5&  62.7 & 27.3 & \bf 54.7  & 39.9\\

\bottomrule
\end{tabular}
}
\caption{Comparison of different class affinity estimation methods with target datasets of 100 images. Results after finetuning.
}
\label{tab:effectClsSimExtrc2}
\end{table*}

\begin{table*}
\centering
 \setlength{\tabcolsep}{5pt} 
\resizebox{\textwidth}{!}
{\tiny
\begin{tabular}{lccccccccc}
\toprule
 & \textbf{Affinity matrix} &  \multicolumn{2}{c}{\textbf{COCO\textrightarrow Cityscapes  }} &  \multicolumn{2}{c}{\textbf{COCO\textrightarrow ADE  }}
  &  \multicolumn{2}{c}{\textbf{ADE\textrightarrow COCO  }}
   &  \multicolumn{2}{c}{\textbf{ADE\textrightarrow Cityscapes  }}
  \\
  &  \textbf{initialization} &
 $\downarrow$\textbf{FID}  &    $\uparrow$\textbf{mIoU}  &
 $\downarrow$\textbf{FID}   &    $\uparrow$\textbf{mIoU} &
 $\downarrow$\textbf{FID}  &    $\uparrow$\textbf{mIoU} &
 $\downarrow$\textbf{FID}  &    $\uparrow$\textbf{mIoU}   \\  
\midrule

 \multirow{6}{*}{\begin{sideways}\textbf{OASIS}\end{sideways}} & Random & 216.4  & 2.0 &216.0 & 0.5 & 270.8  & 0.1 & 326.9 & 2.4  \\

  & Text-based & 130.5 & 26.2    & 44.6  & 23.9  & 57.8  &  15.2  & 138.8  & 35.6 \\
  & Supervised & \textbf{82.1}  & 32.1  & 46.9  & 47.0  & 22.8  & 64.5  & 12.1 & 36.3\\
   & Self-supervised & 88.7 & 31.0  &45.5  & 22.7 & 68.1 & 10.2    &94.2 & 43.3\\
& Combination & 82.4  & 33.4   &  \textbf{43.1}  & \textbf{25.1} & \textbf{56.3}   & \textbf{13.8} & \textbf{79.2}  & 36.9   \\
\midrule
 \multirow{5}{*}{\begin{sideways}\textbf{PITI}\end{sideways}}  &  Random & 254.7 & 3.8 & 96.3 &  5.7 &  98.3&  0.1 &287.4 & 2.0\\ 
  & Text-based & 103.6 & 20.6 & 51.6  & 19.2 & 50.8 &7.1&75.3 &22.3 \\
  & Supervised & \bf 78.1 & 23.6 &  48.7& 20.2 & 59.2 &5.2 & \bf 72.3 &22.9 \\
  & Self-supervised & 88.7 & \bf 24.0 &  49.5 & 19.6  & 53.9 &5.0 &92.9 & 22.1 \\
  & Combination & 78.7 &23.8  & \bf  48.5 &  \bf 20.9 &   \bf 49.3 &   \bf 7.4 & 72.6 & \bf 22.9 \\
\bottomrule
\end{tabular}
}
\caption{Transfer with a target dataset of size 100 using training-free approach.
}
\label{tab:0shotgen2}
\end{table*}

\begin{table*}

\centering
 \setlength{\tabcolsep}{5pt} 
\resizebox{\textwidth}{!}
{\tiny
\begin{tabular}{cccccccccccc}
\toprule
   \multirow{2}{*}{\textbf{FreezeD}}  & \multirow{2}{*}{\textbf{2 stage}} &  \multirow{2}{*}{\textbf{Resid.}}  & \multirow{2}{*}{\textbf{CAT}} &  \multicolumn{2}{c}{\textbf{COCO\textrightarrow ADE  }}
  &  \multicolumn{2}{c}{\textbf{ADE\textrightarrow COCO  }}
   &  \multicolumn{2}{c}{\textbf{ADE\textrightarrow Cityscapes  }} &  \multicolumn{2}{c}{\textbf{COCO\textrightarrow Cityscapes  }}
  \\

& &  & &
 $\downarrow$\textbf{FID}  &    $\uparrow$\textbf{mIoU}  & 
 $\downarrow$\textbf{FID}   &    $\uparrow$\textbf{mIoU}  &
 $\downarrow$\textbf{FID}  &    $\uparrow$\textbf{mIoU}   &
 $\downarrow$\textbf{FID}  &    $\uparrow$\textbf{mIoU}   \\  

\midrule
\xmark & \xmark & \xmark & \xmark & 87.2 & 20.7 &  117.3 & 11.0&56.0&61.6 & 51.3 & 63.7 \\ 
\cmark & \xmark & \xmark & \xmark  &65.7 & 25.8 & 98.6 &14.8&56.9 & 62.9 & 49.7 & 66.6\\ 
\cmark & \cmark & \xmark & \xmark & 55.9 & 28.6 &  83.4 &15.2 & 55.2 & 63.1  & 50.0 &  66.2 \\ 

\cmark & \cmark & \cmark&  \xmark & 55.2 & 29.4 &79.9 & 15.7 & 52.7 & 65.5 &  47.6 & 66.0\\

\cmark & \cmark & \cmark & \cmark   &  \textbf{40.9} &  \textbf{31.4}   &  \textbf{53.7}  & \textbf{17.4}   & \textbf{51.3} &  \textbf{66.4}  &  \textbf{46.9} & \textbf{68.3}  \\
\bottomrule
\end{tabular}
}
\caption{Ablation on OASIS-based architecture.
}
\label{tab:ablationoasis2}
\end{table*}

\begin{table*}

\centering
 \setlength{\tabcolsep}{5pt} 
\resizebox{\textwidth}{!}
{\tiny
\begin{tabular}{ccccccccccccc}
\toprule
    \multirow{2}{*}{\textbf{FixDec}}  & \multirow{2}{*}{\textbf{Prompts}} &  \multirow{2}{*}{\textbf{CAT}}  &  \multicolumn{2}{c}{\textbf{COCO \textrightarrow ADE   }}
  &  \multicolumn{2}{c}{\textbf{ADE  \textrightarrow COCO  }}
   &  \multicolumn{2}{c}{\textbf{ADE  \textrightarrow Cityscapes  }} &  \multicolumn{2}{c}{\textbf{COCO \textrightarrow Cityscapes  }}
  \\

 &  &  & 
 $\downarrow$\textbf{FID}  &    $\uparrow$\textbf{mIoU}  &
 $\downarrow$\textbf{FID}   &    $\uparrow$\textbf{mIoU} &
 $\downarrow$\textbf{FID}  &    $\uparrow$\textbf{mIoU} &
 $\downarrow$\textbf{FID}  &    $\uparrow$\textbf{mIoU}   \\  

\midrule

\xmark & \xmark & \xmark& 84.0 & 0.1 &85.9 & 17.4  &  70.5 & 36.9 & 56.5 &14.4\\
  \cmark & \xmark & \xmark  & 52.4 & 13.6 &  79.0 & 1.4 &66.3 & 20.0&  62.2 & 33.5\\ 
\cmark & \cmark & \xmark  & 51.1 & 14.1 &   78.9 & 1.3&65.4 & 20.3&   58.5 &  33.1\\ 

\cmark & \cmark & \cmark &  \bf 40.7 & \bf 22.3 &  \bf 46.8 & \bf  7.5  & \bf 62.5 & \bf 27.2 & \bf  57.3 & \bf 40.8  \\

\bottomrule
\end{tabular}
}

\caption{Ablation on PITI-based architecture.
}
\label{tab:ablationpiti2}
\end{table*}

\begin{table}
\centering
\setlength{\tabcolsep}{5pt} 
{\small
\begin{tabular}{llcc}
\toprule
  &  &  \multicolumn{1}{c}{\textbf{COCO\textrightarrow ADE  }} &  \multicolumn{1}{c}{\textbf{ADE\textrightarrow COCO  }}
  \\
\midrule
 \multirow{2}{*}{\begin{sideways} \footnotesize  \textbf{OASIS}\end{sideways}}
& Training-free   & 0.011  & 0.025    \\

& After Finetuning   & \bf 0.008 &  \bf 0.024\\

\midrule
 \multirow{2}{*}{\begin{sideways}  \footnotesize  \textbf{PITI}\end{sideways}} 
  & Training-free   &  0.015 & 0.023\\
  & After Finetuning    & \bf 0.010 & \bf 0.020 \\
\bottomrule
\end{tabular}
}
\caption{KID comparing training-free and finetuning mode
}
\label{tab:KID}
\end{table}

\section{Additional qualitative results}
\label{app:qualitative}

\mypar{Comparison with the state of the art} 
In Figure~\ref{fig:qualBaselinesOASIS} and Figure~\ref{fig:qualBaselinesPITI}, we compare our CAT approach to the best state-of-the-art models in Table 2 of the main paper. 
For OASIS, we compare to cGANTransfer~\cite{shahbazi21cvpr}, TransferGAN~\cite{wang18eccv2}, FreezeD~\cite{mo20cvpr} as they get the best FID/mIoU scores for one of the four source-target dataset pairs. We show samples for the four source-target dataset pairs and observe images with superior quality using our approach. 
It is particularly striking for PITI when comparing to  finetuning all  layers from a pretrained PITI model,  or finetuning from a GLIDE checkpoint. In this case, generated images do not adhere well to label maps and are of poor quality.

\mypar{Qualitatives with different target dataset sizes} In Figure~\ref{fig:diffDataSize}, we complete Figure 4 of main paper by showing qualitative samples from OASIS with COCO-Stuff as  source and using  ADE20K as target,  with target dataset of sizes 25, 100, 400, 
 and 20k, where 20k corresponds to the full dataset. 
 We  notice how image quality gradually improves with the number of images in the target dataset.

\mypar{Training-free samples}
In Figure \ref{fig:trainingfree}, we show samples obtained without finetuning the model, by using  the pretrained weights and only adding our   affinity matrix  mapping to the model.
We compare them to synthesized images conditioned on the same segmentation map and noise input after finetuning on three pairs of source-target datasets:  COCO-Stuff\textrightarrow ADE20K and vice-versa, as well as  COCO-Stuff\textrightarrow Cityscapes. 

We  observe that target classes which do not exist in source dataset are poorly rendered in the training-free mode. 
For instance, the closest COCO-Stuff class to ``painting'' in ADE20K is ``building-other'' according to our combination method, while ``cradle'' is associated to COCO-Stuff class ``bed''. 
We can see in the top right block of the Figure~\ref{fig:trainingfree} in rows 2 and 4 that these objects are better recognizable after finetuning. 
It is also worth noticing that both PITI and OASIS generators adapt well to the style of the target dataset when transferring from COCO-Stuff or ADE20K to Cityscapes. While training-free samples look colorful and bright, images after finetuning are darker, more in line with the style of the original Cityscapes images. 

\input{FigBaselineQual}
\input{figBaselinePitiQual}
\input{diffDataSizeFig}

\input{FigTrainFree}

%% file: hardsoft.tex
\mypar{Hard or soft affinity matrices}
In Table \ref{tab:softBinAffMat}, we conduct experiments by using hard or soft affinity matrices. 
More precisely, we compute the soft affinity matrices as described in Section 3.1 of the main paper.
For the ``hard'' version,  for each target class we binarize the affinities by setting the largest value to one, and the rest to zeros. 
The results show that  the hard version  yields consistent gains in FID and mIoU for both for PITI and OASIS, and we retain it in our main experiments.

\begin{table}
\centering
 \setlength{\tabcolsep}{5pt} 
{\small
\begin{tabular}{lccccc}
\toprule
  \multirow{2}{*}{\textbf{Model}}&\multirow{2}{*}{\textbf{Aff.\ mat.\ }} &   \multicolumn{2}{c}{\textbf{COCO $\rightarrow$ ADE  }}  \\
  & 
& $\downarrow$\textbf{FID}  &    $\uparrow$\textbf{mIoU}   \\  
\midrule
 \multirow{2}{*}{\textbf{OASIS}} & Hard & \bf 40.9 & \bf 31.4 \\
& Soft  & 44.7 & 30.1\\
\midrule
\multirow{2}{*}{\textbf{PITI}} & Hard & \bf 40.7 & \bf 22.3  \\
& Soft & 43.6 & 19.8 \\
\bottomrule
\end{tabular}
}
\caption{Ablation on the use of hard \vs soft affinity matrices. Affinities computed with  supervised segmentation network. \label{tab:softBinAffMat} 
}
\end{table}

%% file: FigBaselineQual.tex
\begin{figure*}
 \def\myim#1{ \includegraphics[width=17.0mm,height=17.0mm]{samples/#1}}
 \def\myimCS#1{ \includegraphics[width=34.5mm,height=17.25mm]{samples/#1}}
     \centering
   \setlength\tabcolsep{0.5 pt}
   \renewcommand{\arraystretch}{0.2}
   \footnotesize
     \begin{tabular}{cccccccccccc}
Input & TransferGAN  & FreezeD & cGANTransfer  & CAT (ours)&\hspace{0.1cm}&Input & TransferGAN  & FreezeD & cGANTransfer  & CAT (ours)\\

\myim{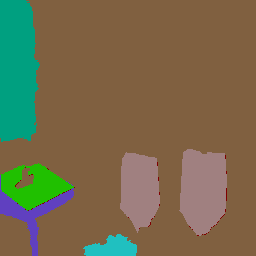} &
\myim{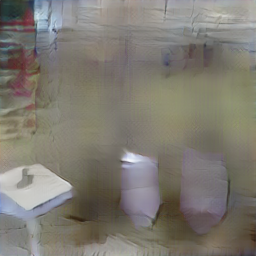} &
\myim{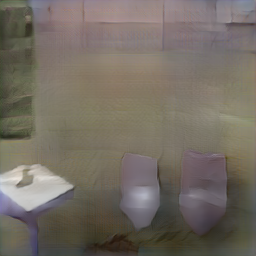} &
\myim{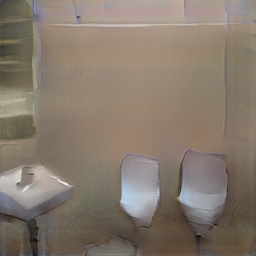} &
\myim{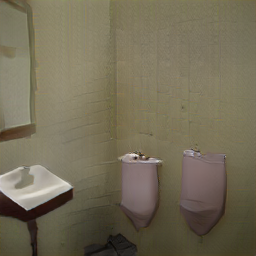} 

&\hspace{0.1cm}&
\myim{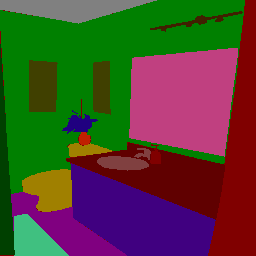} &
\myim{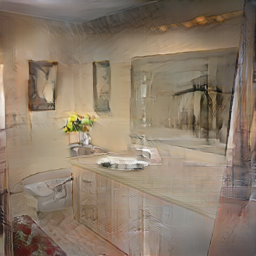} &
\myim{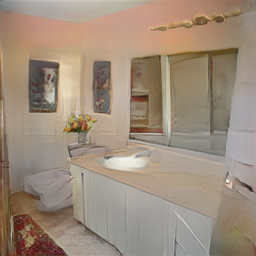} &
\myim{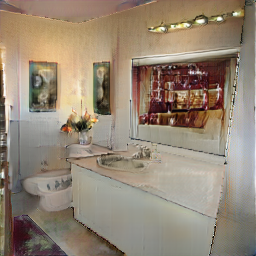} &
\myim{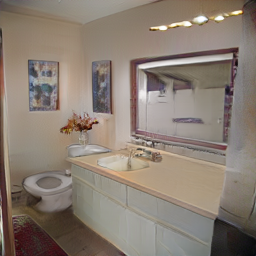}  \\

\myim{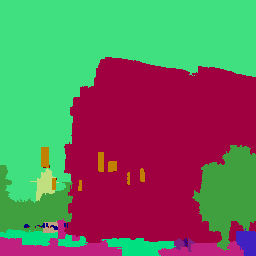} &
\myim{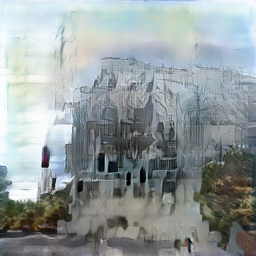} &
\myim{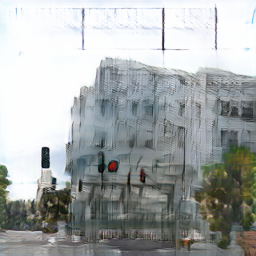} &
\myim{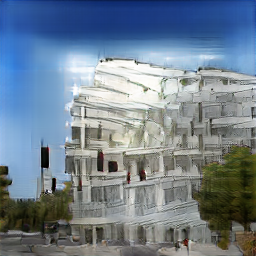} &
\myim{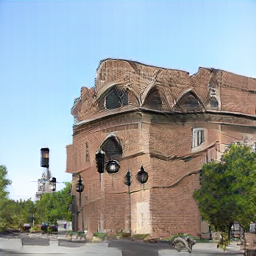} 

&\hspace{0.1cm} &
\myim{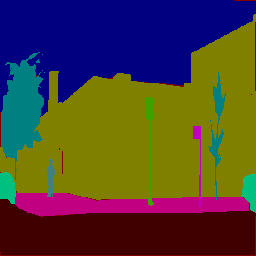} &
\myim{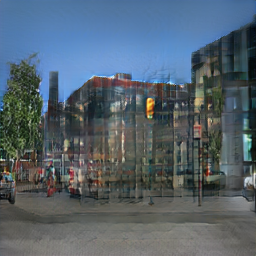} &
\myim{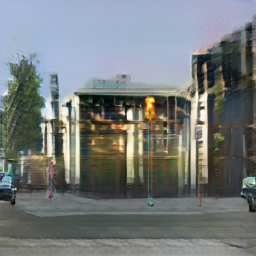} &
\myim{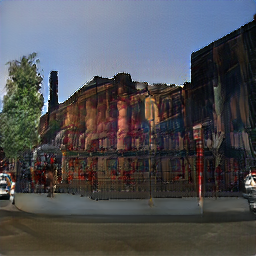} &
\myim{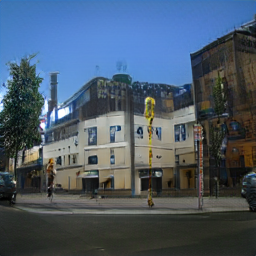} \\

\myim{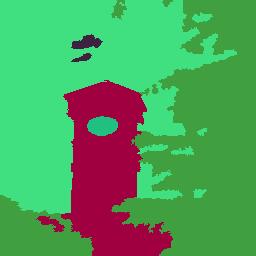} &
\myim{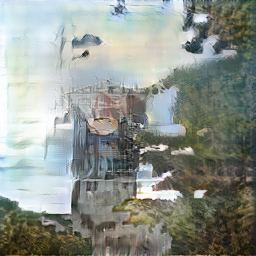} &
\myim{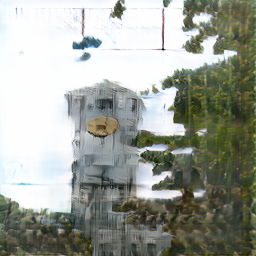} &
\myim{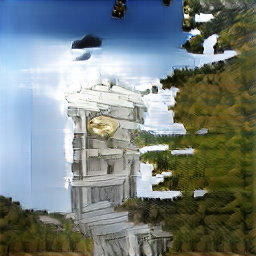} &
\myim{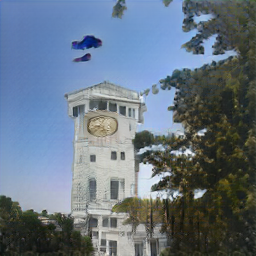} 
&\hspace{0.1cm}&
\myim{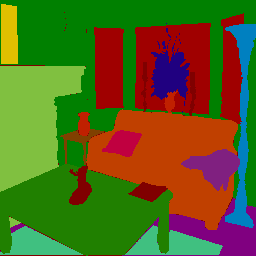} &
\myim{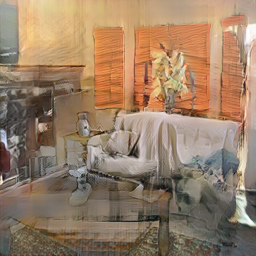} &
\myim{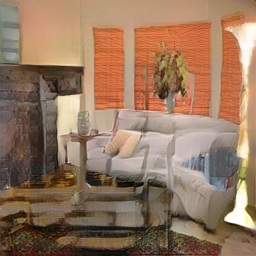} &
\myim{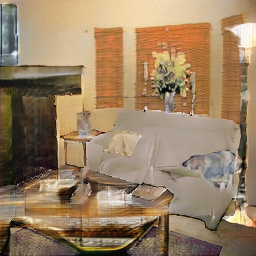} &
\myim{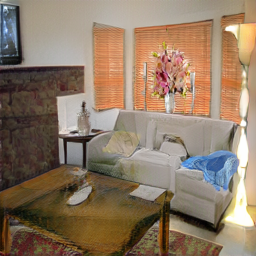} \\

\myim{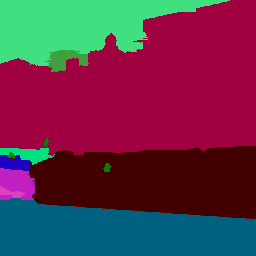} &
\myim{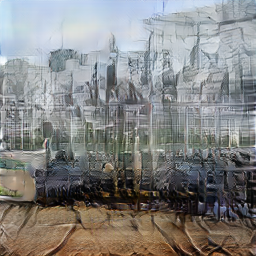} &
\myim{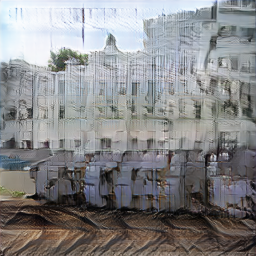} &
\myim{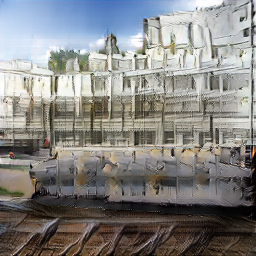} &
\myim{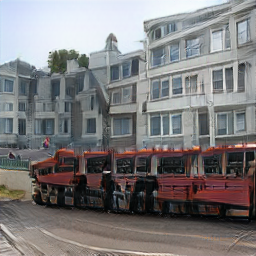} 
&\hspace{0.1cm}&
\myim{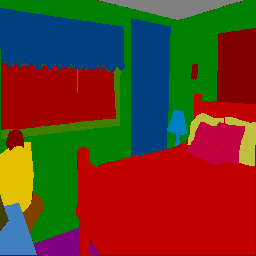} &
\myim{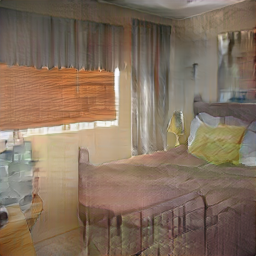} &
\myim{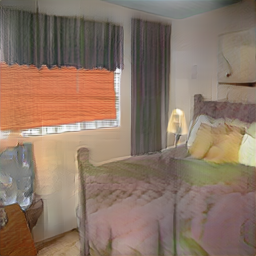} &
\myim{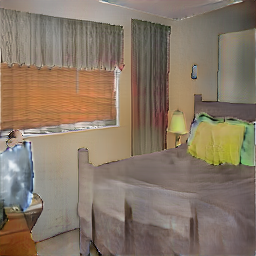} &
\myim{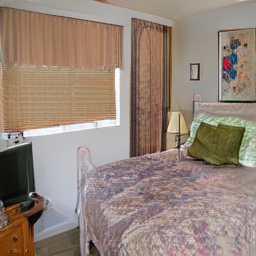}  \\

\end{tabular}
\\
\vspace{5mm}

\begin{tabular}{ccccc}
\\
Input & TransferGAN  & FreezeD & cGANTransfer  & CAT (ours)\\
\myimCS{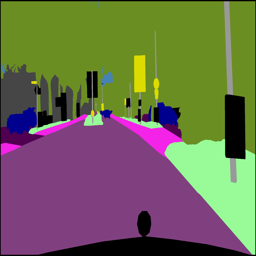} &
\myimCS{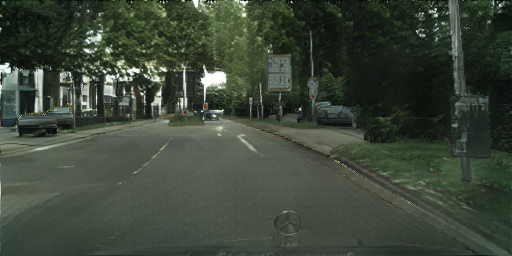} & 
\myimCS{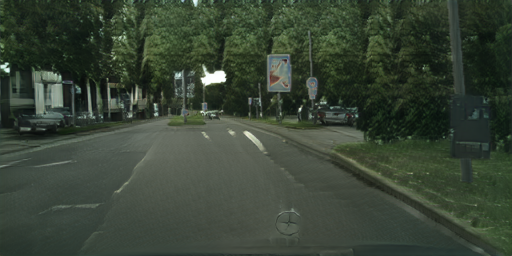}&
\myimCS{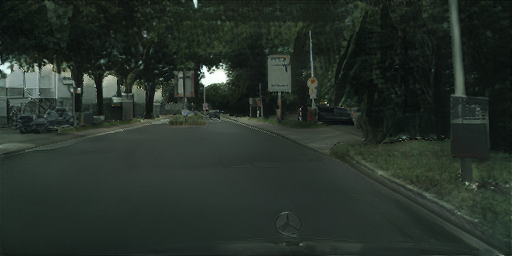} & 
\myimCS{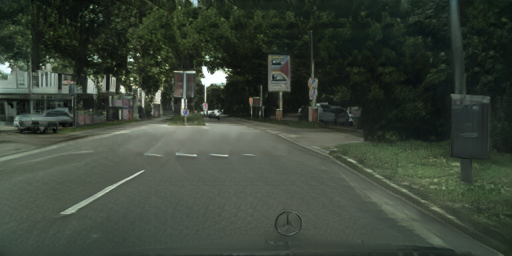}\\

\myimCS{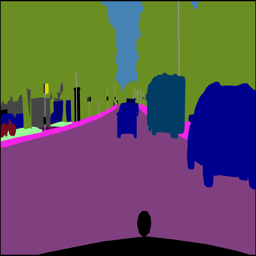} &
\myimCS{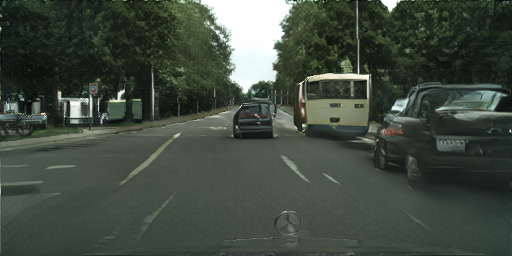} & 
\myimCS{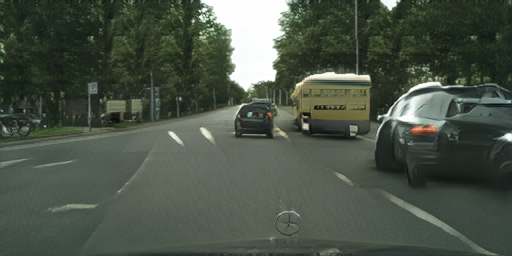}&
\myimCS{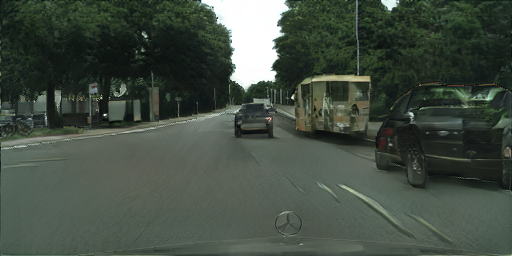} & 
\myimCS{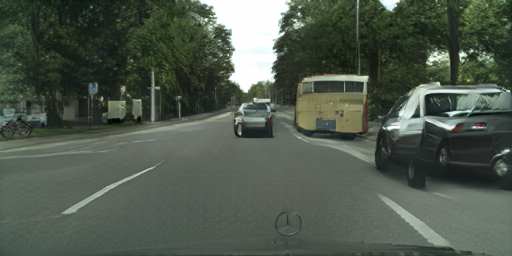}\\

\myimCS{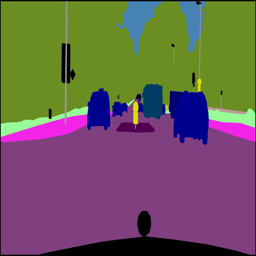} &
\myimCS{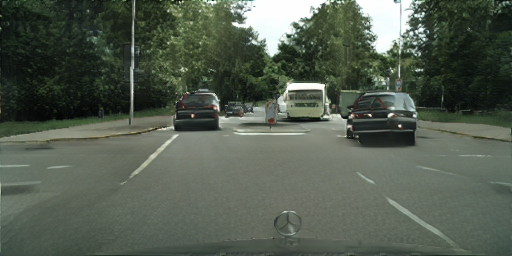} & 
\myimCS{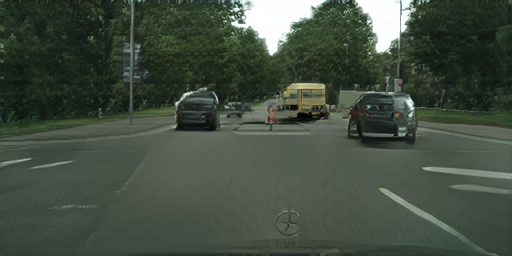}&
\myimCS{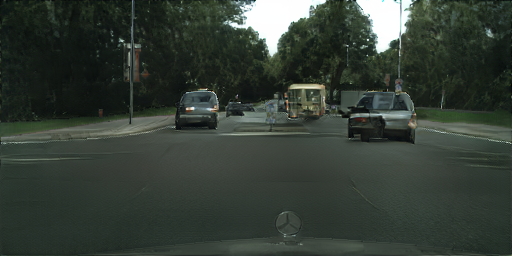} & 
\myimCS{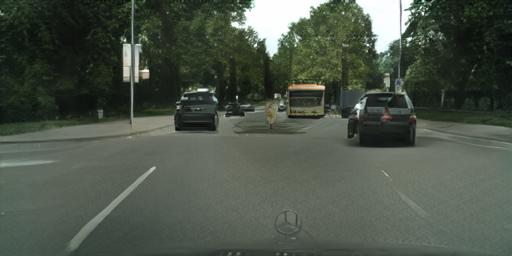}\\
\vspace{0.5cm} \\

Input & TransferGAN  & FreezeD & cGANTransfer  & CAT (ours)\\
\myimCS{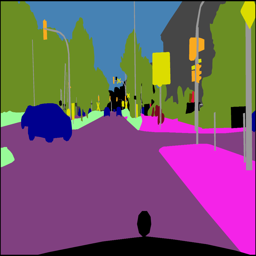} &
\myimCS{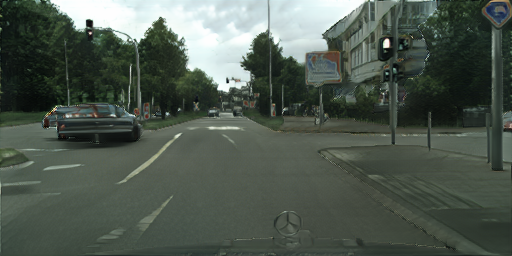} & 
\myimCS{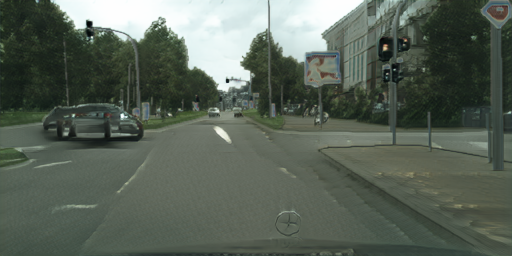}&
\myimCS{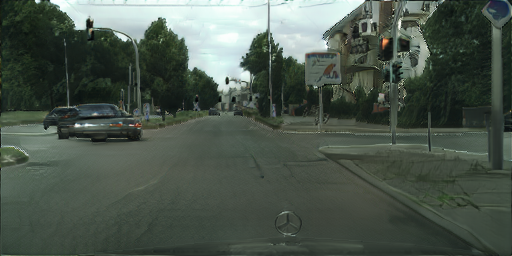} & 
\myimCS{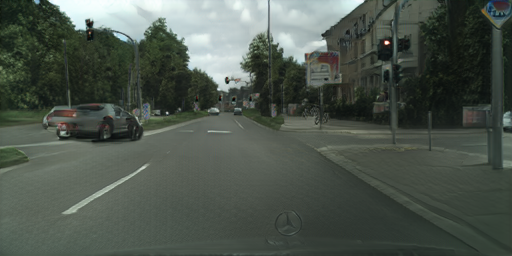}\\

\myimCS{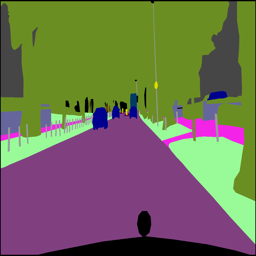} &
\myimCS{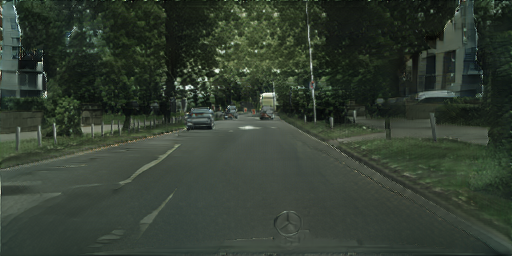} & 
\myimCS{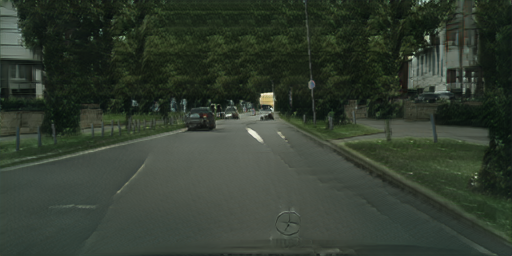}&
\myimCS{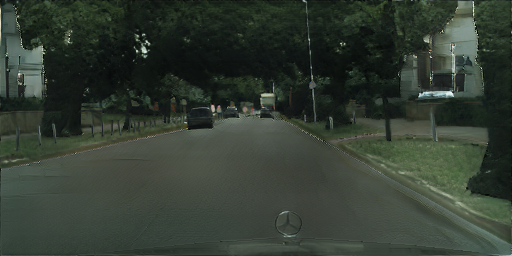} & 
\myimCS{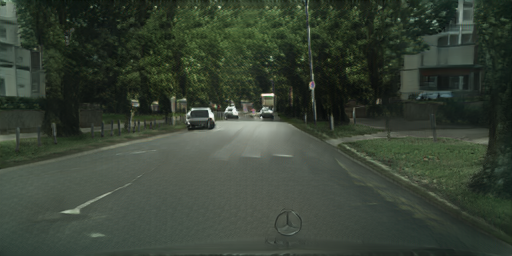}\\

\myimCS{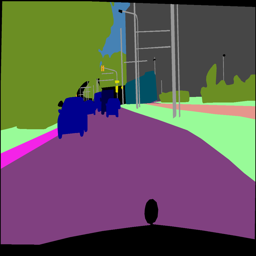} &
\myimCS{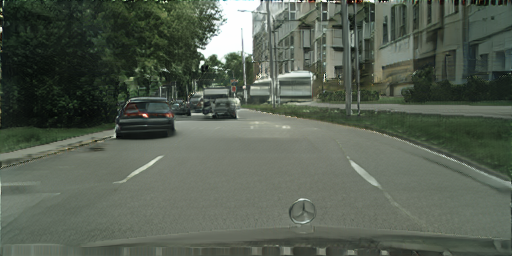} & 
\myimCS{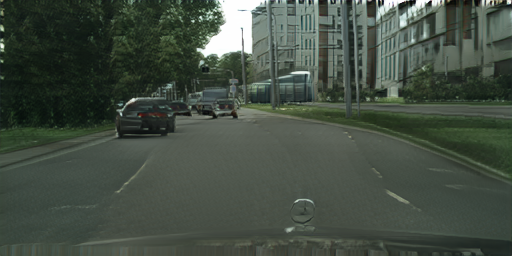}&
\myimCS{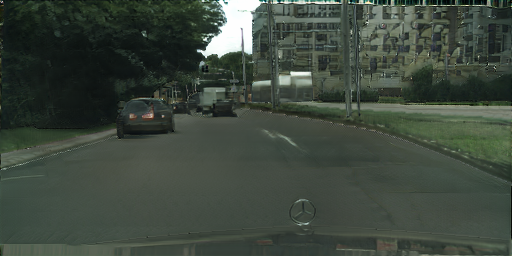} & 
\myimCS{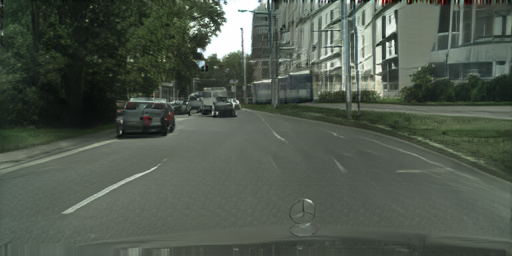}\\

\end{tabular}

\caption{Samples from OASIS finetuned with 100 target images.
Transfer from ADE to COCO (top left), COCO to ADE (top right), from COCO to Cityscapes (middle) and from ADE to Cityscapes (bottom).
}
\label{fig:qualBaselinesOASIS}
\end{figure*}

%% file: figBaselinePitiQual.tex
\begin{figure*}
 \def\myim#1{ \includegraphics[width=17.0mm,height=17.0mm]{samples/#1}}
 \def\myimCS#1{ \includegraphics[width=34.5mm,height=17.25mm]{samples/#1}}
     \centering
   \setlength\tabcolsep{0.5 pt}
   \renewcommand{\arraystretch}{0.2}
   \footnotesize
     \begin{tabular}{cccccccccc}
\\
Input & From GLIDE  & Finetune all &  CAT (ours)&\hspace{0.1cm}&Input & From GLIDE  & Finetune all &  CAT (ours)\\

\myim{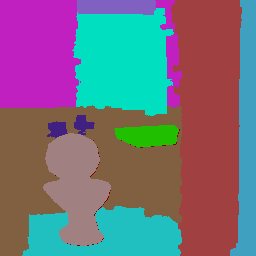} &
\myim{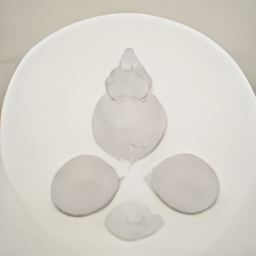} &
\myim{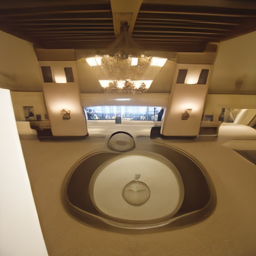} &
\myim{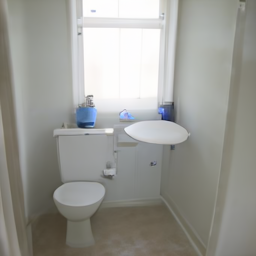}

&\hspace{0.1cm}&
\myim{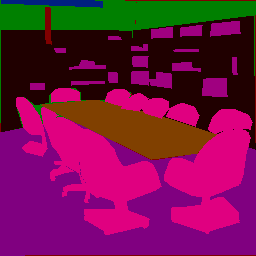} &
\myim{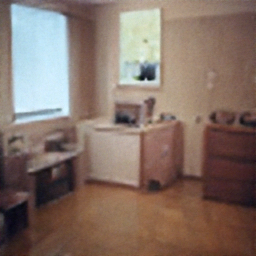} &
\myim{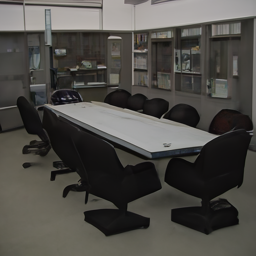} &
\myim{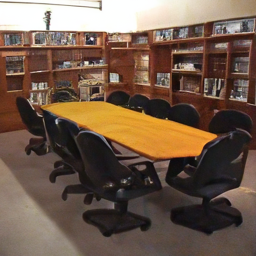}  \\

\myim{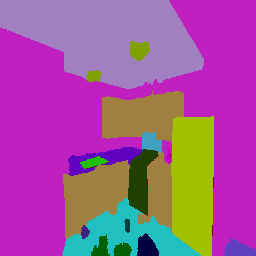} &
\myim{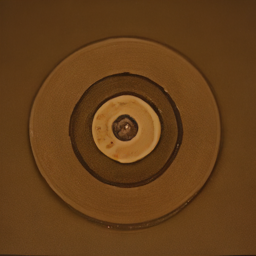} &
\myim{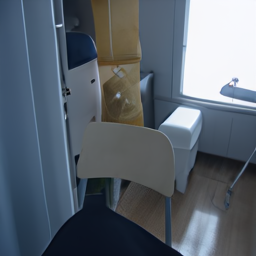} &
\myim{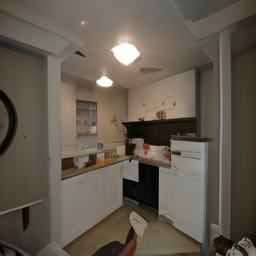} 

&\hspace{0.1cm} &
\myim{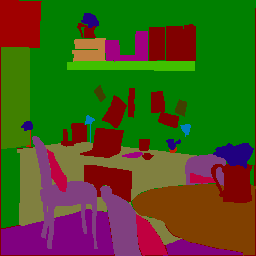} &
\myim{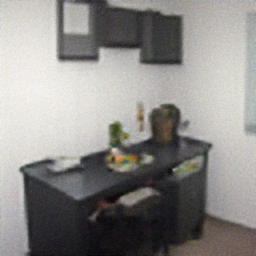} &
\myim{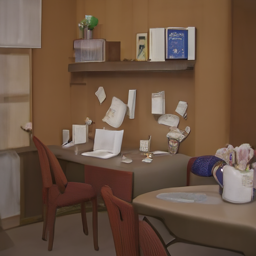} &
\myim{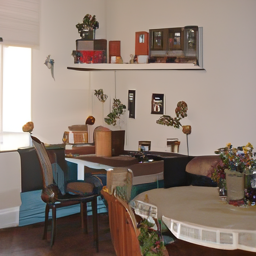}  \\

\myim{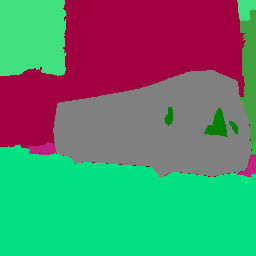} &
\myim{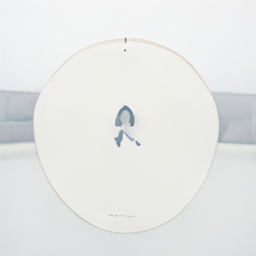} &
\myim{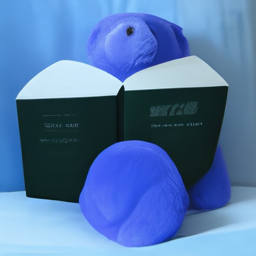} &
\myim{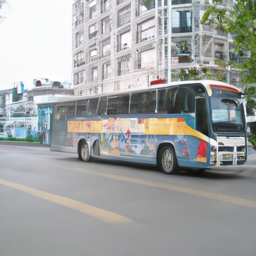} 

&\hspace{0.1cm}&
\myim{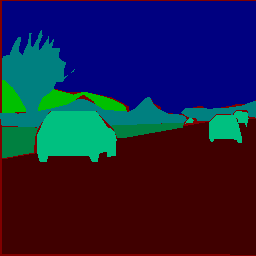} &
\myim{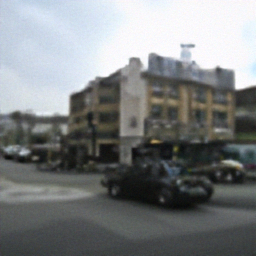} &
\myim{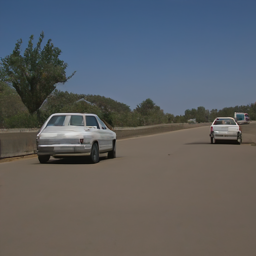} &
\myim{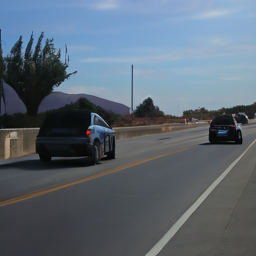}  \\

\end{tabular}
\\
\vspace{5mm}

\begin{tabular}{cccc}

\\
Input & From GLIDE  & finetune all & CAT (ours)\\
\myimCS{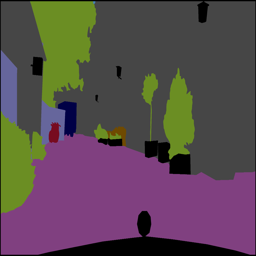} &
\myimCS{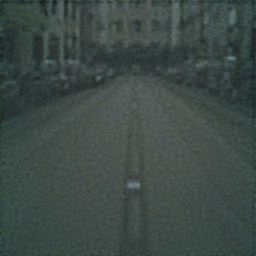} & 
\myimCS{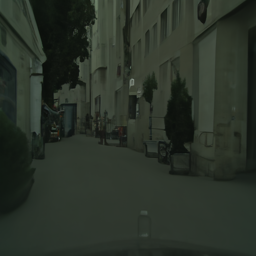}&
\myimCS{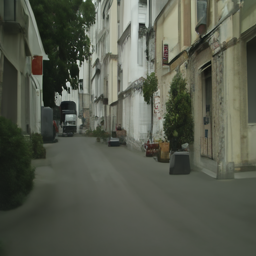}\\

\myimCS{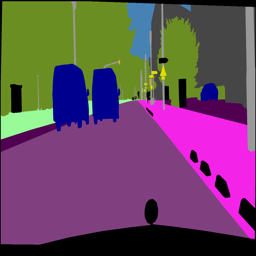} &
\myimCS{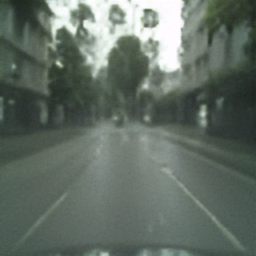} & 
\myimCS{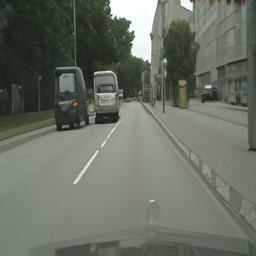}&
\myimCS{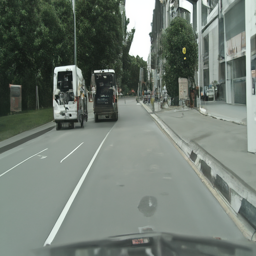}\\

\myimCS{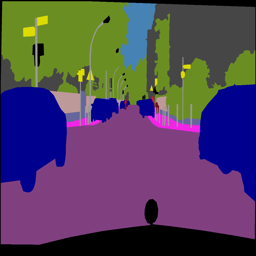} &
\myimCS{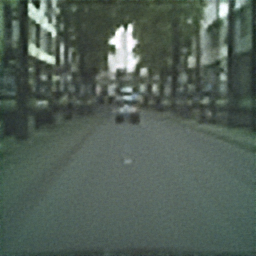} & 
\myimCS{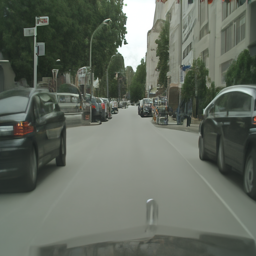}&
\myimCS{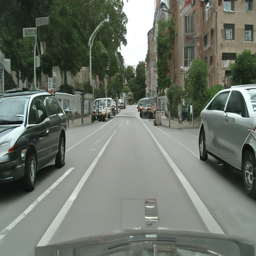}\\

\vspace{0.5cm}\\
Input & From GLIDE  & finetune all & CAT (ours) \\
\myimCS{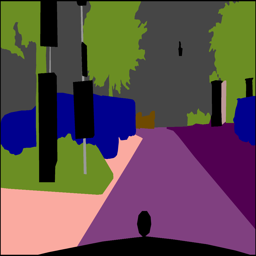} &
\myimCS{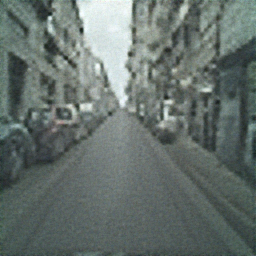} & 
\myimCS{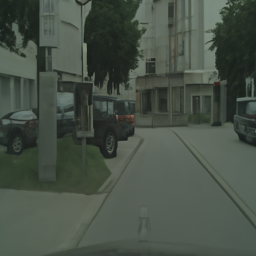}&
\myimCS{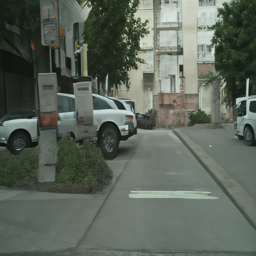}\\

\myimCS{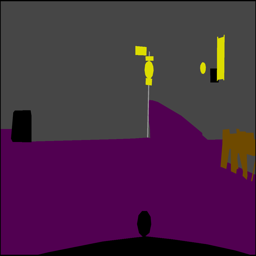} &
\myimCS{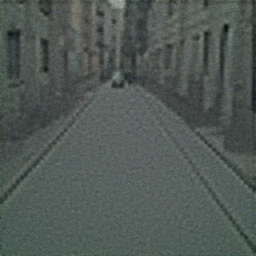} & 
\myimCS{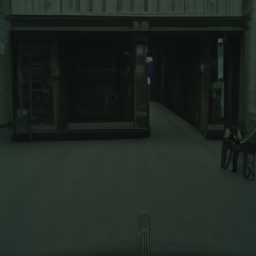}&
\myimCS{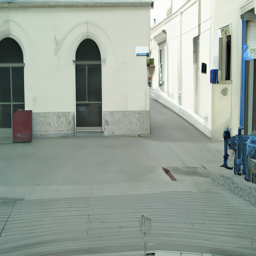}\\

\myimCS{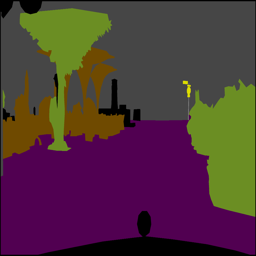} &
\myimCS{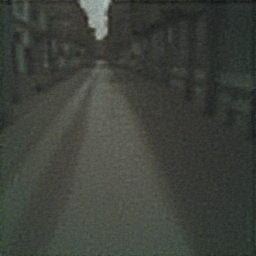} & 
\myimCS{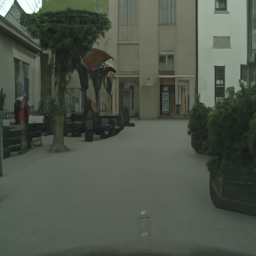}&
\myimCS{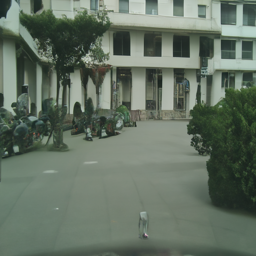}\\

\end{tabular}

\caption{Samples from PITI finetuned with 100 target images.
Transfer from ADE to COCO (top left), COCO to ADE (top right), from COCO to Cityscapes (middle) and from ADE to Cityscapes (bottom). 
}
\label{fig:qualBaselinesPITI}
\end{figure*}

%% file: diffDataSizeFig.tex
\begin{figure*}
 \def\myim#1{\includegraphics[width=20mm,height=20mm]{samples/ade20k_100_oasis_main/diffDataSize/#1}}
 \def\myimC#1{\includegraphics[width=17mm,height=17mm]{samples/ade20k_100_oasis_main/#1}}
     \centering
   \setlength\tabcolsep{1.5 pt}
   \small
     \begin{tabular}{ccccc}
     & \multicolumn{4}{c}{Size of target training set} \\
Input segmentation & 25  & 100 & 400& 20k (Full dataset) \\
\myim{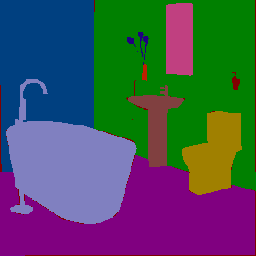} &
\myim{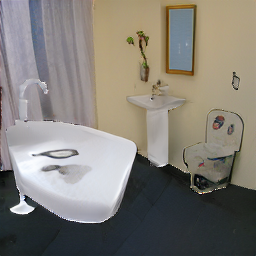} &
\myim{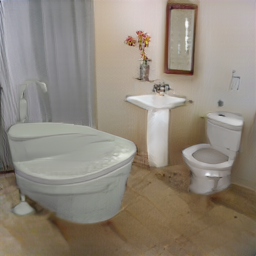} & 
\myim{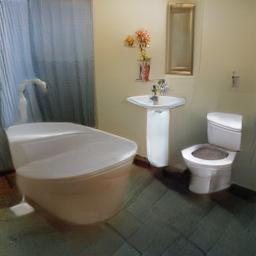} &
\myim{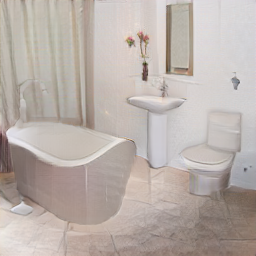} 
\\
\myim{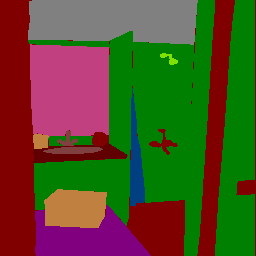} &
\myim{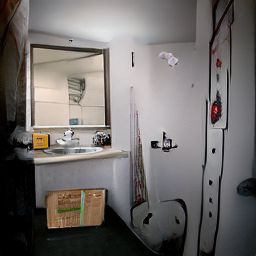} &
\myim{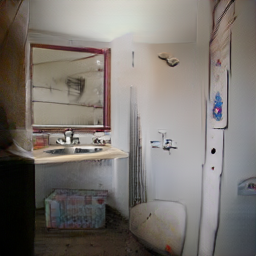} & 
\myim{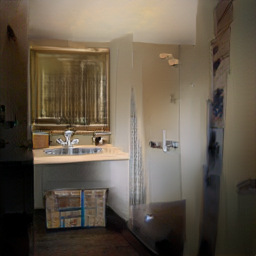} &
\myim{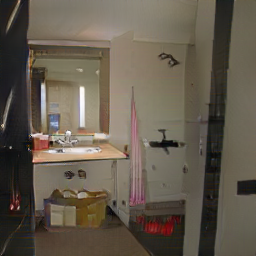} 
\\
\myim{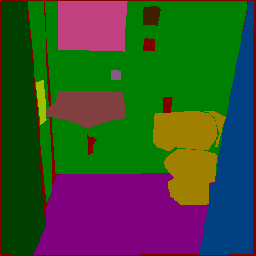} &
\myim{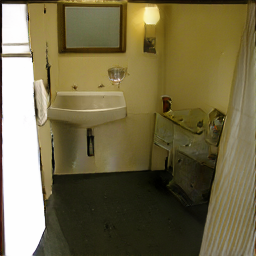} &
\myim{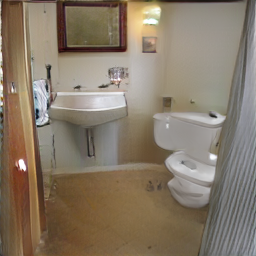} & 
\myim{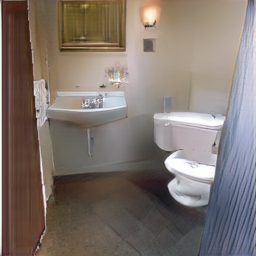} &
\myim{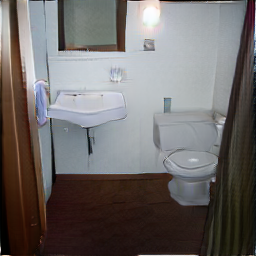} 
\\
\myim{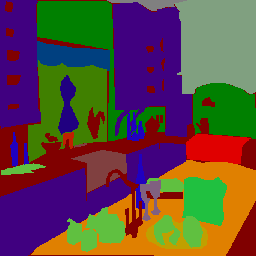} &
\myim{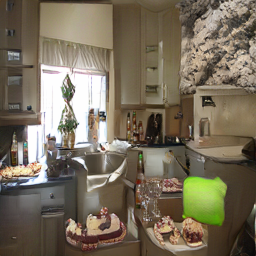} &
\myim{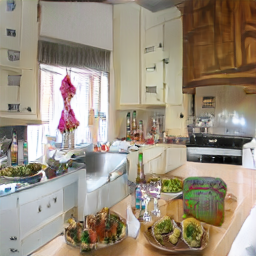} & 
\myim{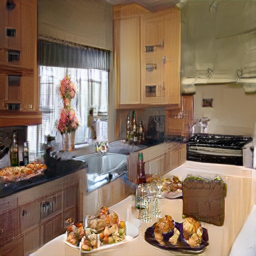} &
\myim{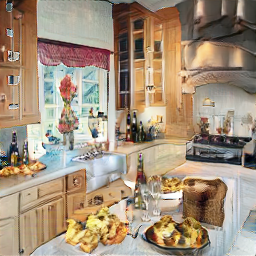} 
\\
\myim{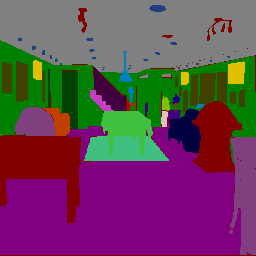} &
\myim{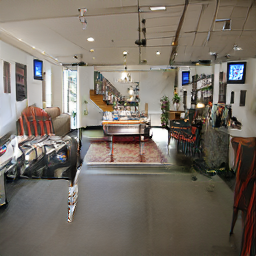} &
\myim{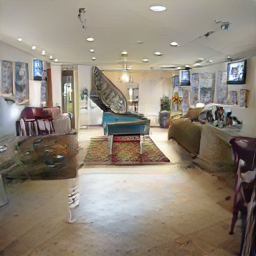} & 
\myim{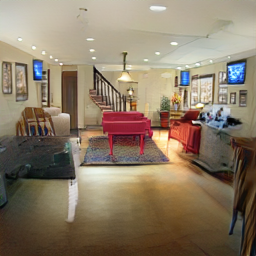} &
\myim{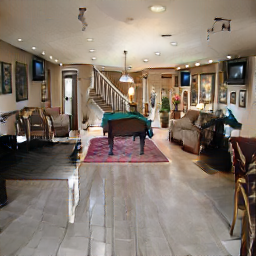} 
\\
\myimC{gt_seg/ADE_val_00001837.png} &
\myim{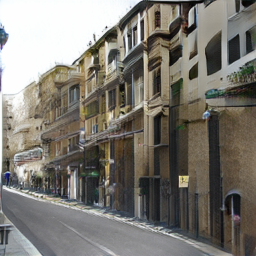} &
\myimC{wConfMat/ADE_val_00001837.png} & 
\myim{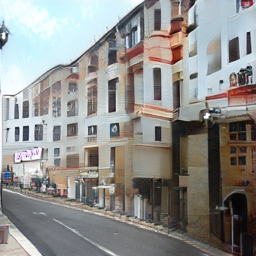} &
\myim{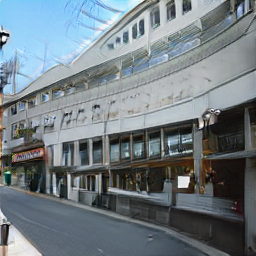} 
\\
\end{tabular}
\caption{ Samples from OASIS finetuned with target datasets from size in \{25,100,400,20k\}. Transfer from COCO to ADE.}
\label{fig:diffDataSize}
\end{figure*}

%% file: FigTrainFree.tex
\begin{figure*}
 \def\myim#1{ \includegraphics[width=17.0mm,height=17.0mm]{samples/#1}}
 \def\myimCS#1{ \includegraphics[width=34.5mm,height=17.25mm]{samples/#1}}
     \centering
   \setlength\tabcolsep{0.5 pt}
   \renewcommand{\arraystretch}{0.2}
   \footnotesize
     \begin{tabular}{cccccccccccc}
&   \multicolumn{2}{c}{OASIS}&   \multicolumn{2}{c}{PITI}&\hspace{0.1cm}&&   \multicolumn{2}{c}{OASIS}&   \multicolumn{2}{c}{PITI}\\
\\
Input & Training-free  & finetuning & Training-free  & finetuning &\hspace{0.1cm}&Input & Training-free  & finetuning & Training-free  & finetuning\\

\myim{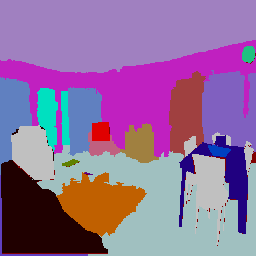} &
\myim{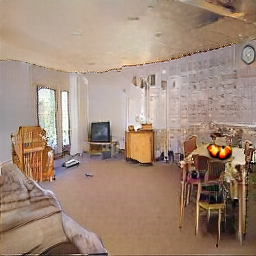} &
\myim{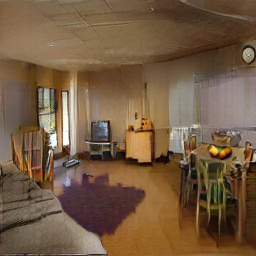} &
\myim{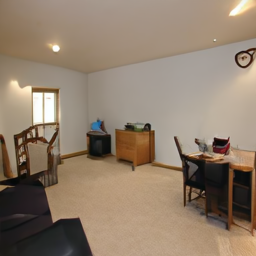} &
\myim{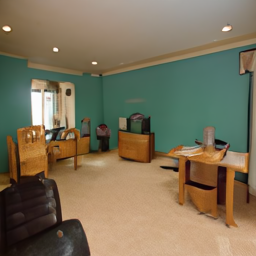} 
&\hspace{0.1cm}&
\myim{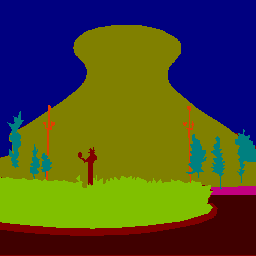} &
\myim{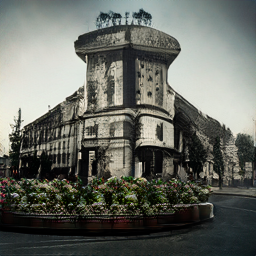} &
\myim{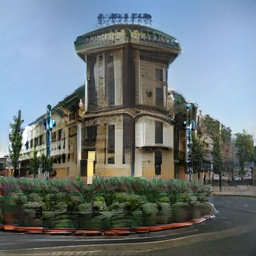}&
\myim{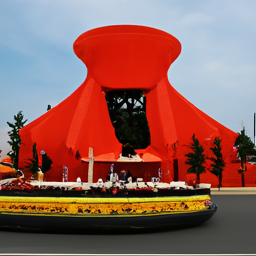} &
\myim{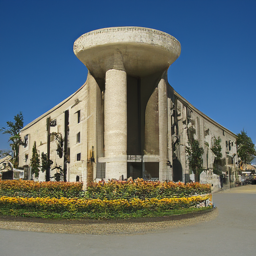} \\

\myim{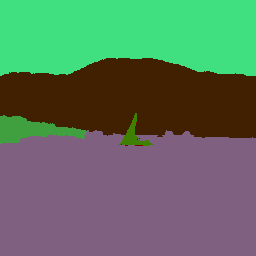} &
\myim{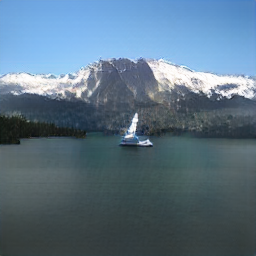} &
\myim{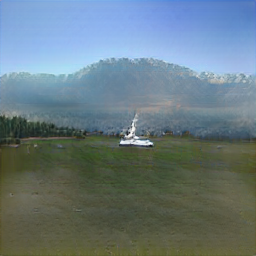} &
\myim{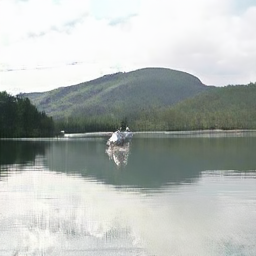} &
\myim{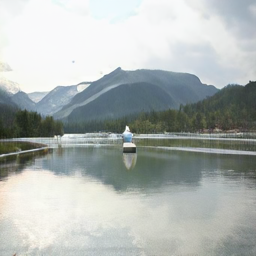} 
&\hspace{0.1cm} &
\myim{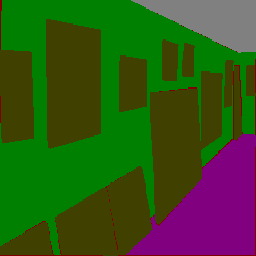} &
\myim{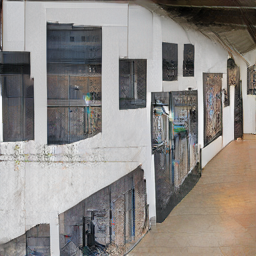} &
\myim{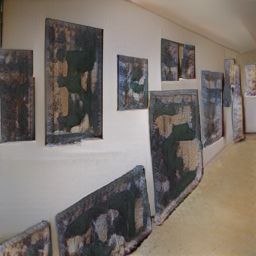} &
\myim{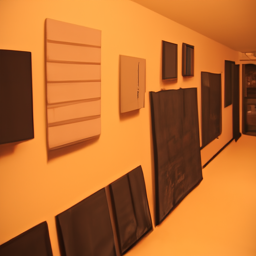} &
\myim{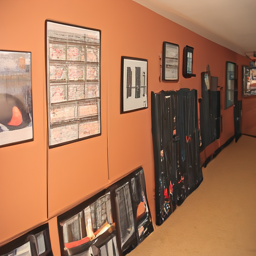} &\\

\myim{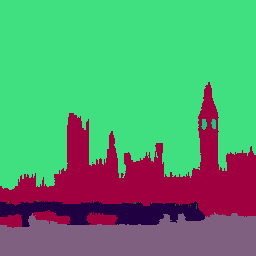} &
\myim{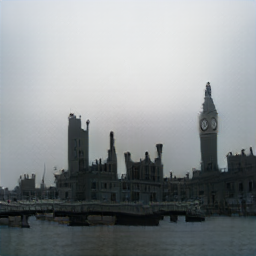} &
\myim{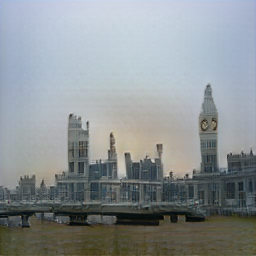} &
\myim{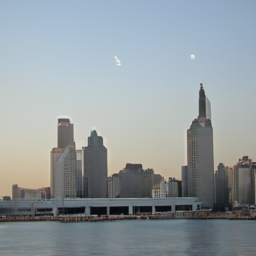} &
\myim{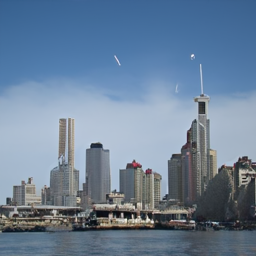} 
&\hspace{0.1cm}&
\myim{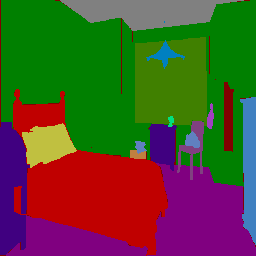} &
\myim{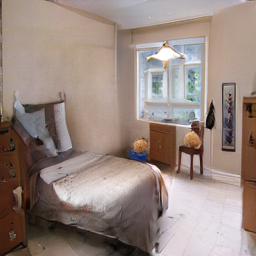} &
\myim{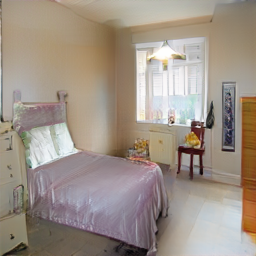}&
\myim{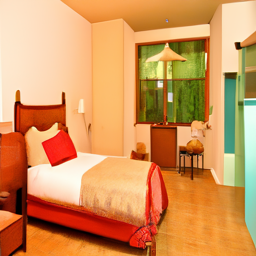} &
\myim{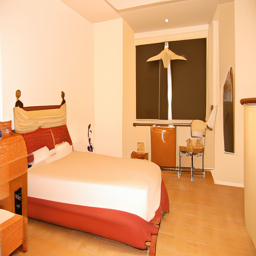} \\

\myim{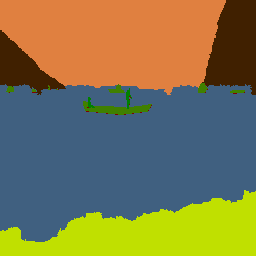} &
\myim{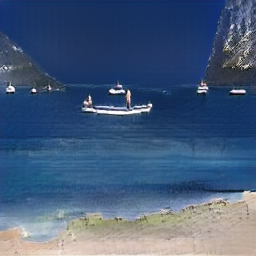} &
\myim{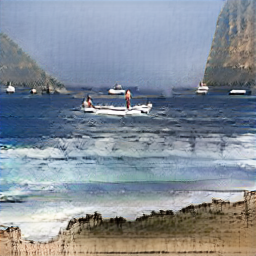} &
\myim{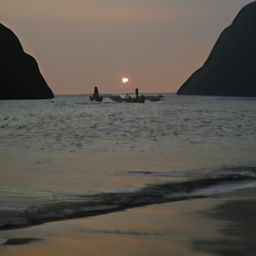} &
\myim{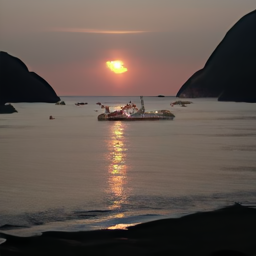} 
&\hspace{0.1cm}&
\myim{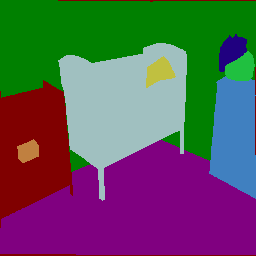} &
\myim{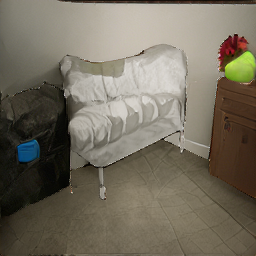} &
\myim{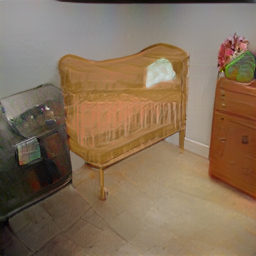} &
\myim{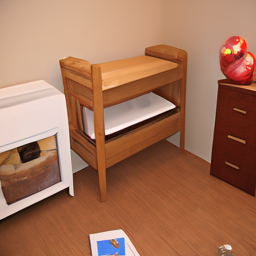} &
\myim{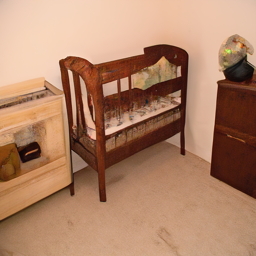} 
\end{tabular}
\\
\vspace{5mm}

\begin{tabular}{ccccc}
&   \multicolumn{2}{c}{OASIS}&   \multicolumn{2}{c}{PITI}\\
\\
Input & Training-free  & finetuning & Training-free  & finetuning\\
\myimCS{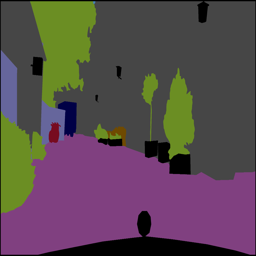} &
\myimCS{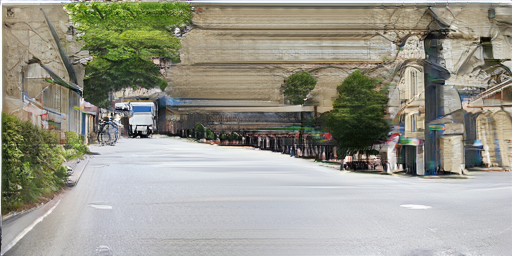} & 
\myimCS{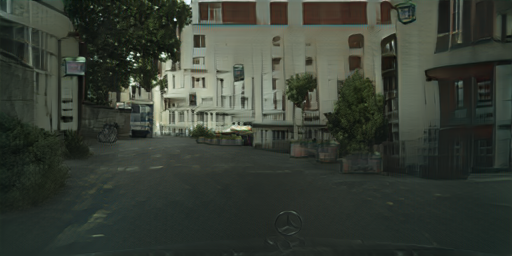}&
\myimCS{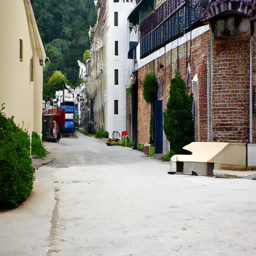} & 
\myimCS{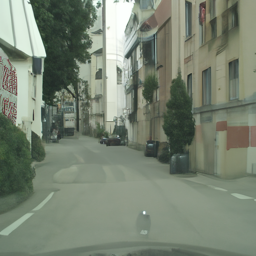}\\

\myimCS{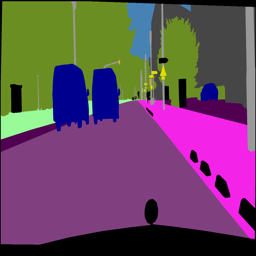}&
\myimCS{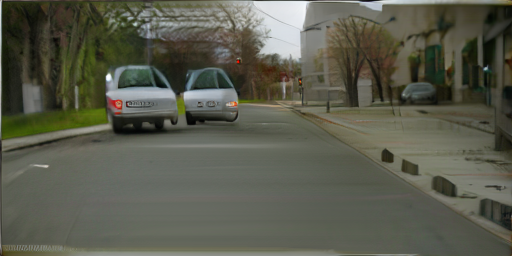} &
\myimCS{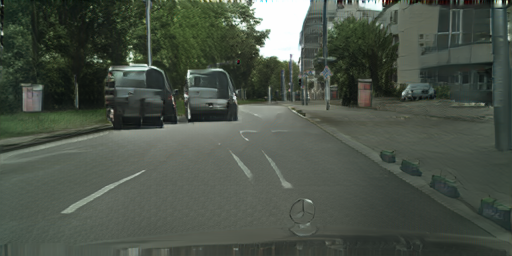}&
\myimCS{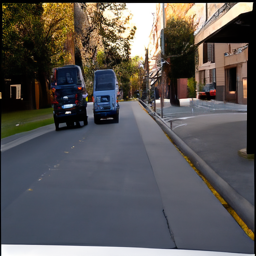} &
\myimCS{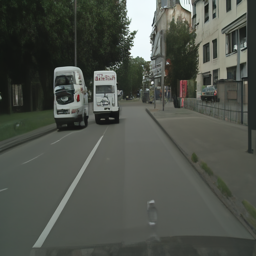}\\

\myimCS{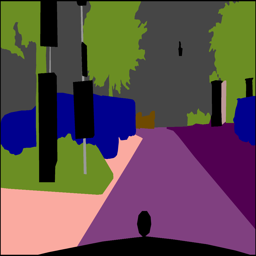}&
\myimCS{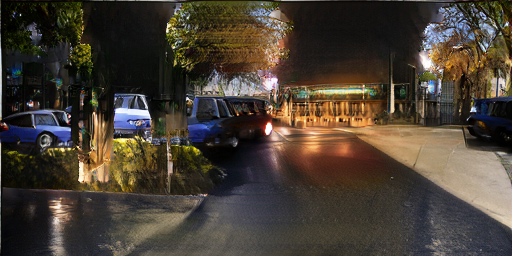} &
\myimCS{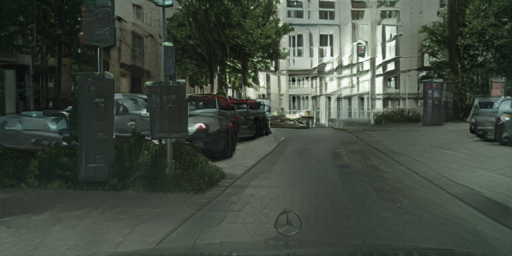}&
\myimCS{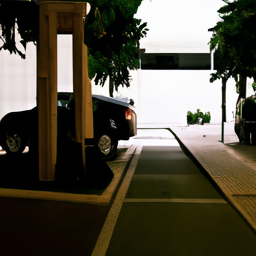} &
\myimCS{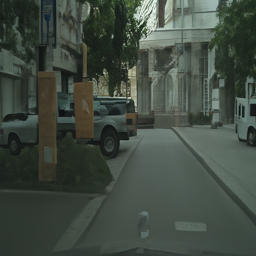}\\

\myimCS{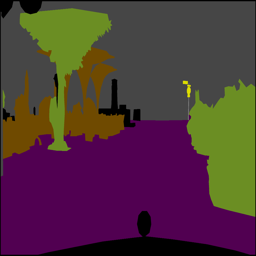}&
\myimCS{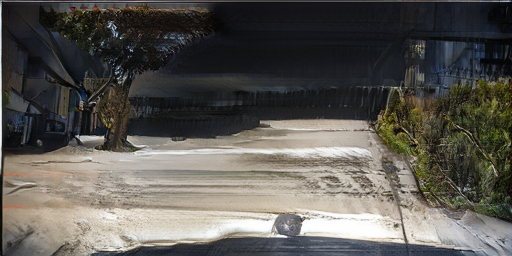} &
\myimCS{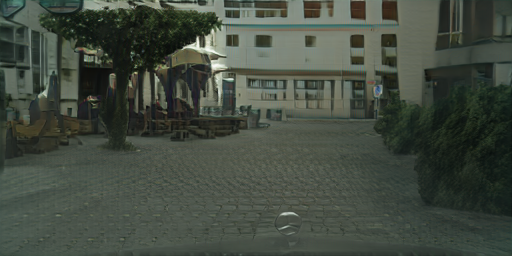}&
\myimCS{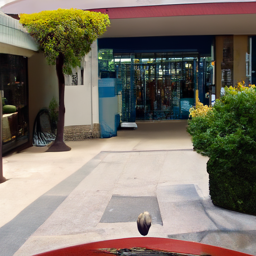} &
\myimCS{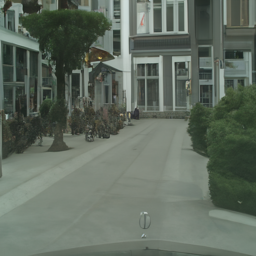}
\end{tabular}
\caption{Samples from models trained with 100 target images.
Transfer from ADE to COCO (top left), COCO to ADE (top right), and from COCO to Cityscapes (bottom). Samples obtained in training-free mode and after finetuning. Samples conditioned on the same noise.
}
\label{fig:trainingfree}
\end{figure*}